\documentclass[manuscript,screen]{acmart}
\usepackage{graphicx}
\graphicspath{{figures/}}
\usepackage{amsmath}
\usepackage{diagbox}

\usepackage{amssymb}
\usepackage{booktabs}
\usepackage{subcaption}
\usepackage{multirow}
\usepackage{pifont}   
\usepackage{bbding} 
\usepackage{fontawesome} 
\usepackage[linesnumbered,ruled]{algorithm2e}
\usepackage{array}
\usepackage{fixltx2e}
\usepackage{stfloats}
\usepackage{caption}
\usepackage{threeparttable}

\AtBeginDocument{%
  \providecommand\BibTeX{{%
    \normalfont B\kern-0.5em{\scshape i\kern-0.25em b}\kern-0.8em\TeX}}}

\setcopyright{acmcopyright}
\copyrightyear{2022}
\acmYear{2022}
\acmDOI{XXXXXXX.XXXXXXX}

\begin{document}

\title{EViT: Privacy-Preserving Image Retrieval via Encrypted Vision Transformer in Cloud Computing}

\author{Qihua Feng}
\email{fengqh2020@gmail.com}
\affiliation{%
  \institution{Jinan University}
  \country{Guang Zhou, China}
  \postcode{510000}
}

\author{Peiya Li}
\authornote{Corresponding author.}
\email{lpy0303@jnu.edu.cn}
\affiliation{%
  \institution{Jinan University}
  \country{Guang Zhou, China}
  \postcode{510000}
}

\author{Zhixun Lu}
\affiliation{%
  \institution{Jinan University}
  \country{Guang Zhou, China}
  \postcode{510000}
}

\author{Chaozhuo Li}
\affiliation{%
  \institution{Beihang University}
  \country{Beijing, China}
  \postcode{100000}
}

\author{Zefang Wang, Zhiquan Liu}
\affiliation{%
  \institution{Jinan University}
  \country{Guang Zhou, China}
  \postcode{510000}}

\author{Chunhui Duan}
\affiliation{%
  \institution{Beijing Institute of Technology}
  \country{Beijing, China}
  \postcode{100000}}

\author{Feiran Huang}
\affiliation{%
  \institution{Jinan University}
  \country{Guang Zhou, China}
  \postcode{510000}}
\renewcommand{\shortauthors}{Qihua Feng, et al.}

\begin{abstract}
  Image retrieval systems help users to browse and search among extensive images in real-time. With the rise of cloud computing, retrieval tasks are usually outsourced to cloud servers. However, the cloud scenario brings a daunting challenge of privacy protection as cloud servers cannot be fully trusted. To this end, image-encryption-based privacy-preserving image retrieval schemes have been developed, which first extract features from cipher-images, and then build retrieval models based on these features. Yet, most existing approaches extract shallow features and design trivial retrieval models, resulting in insufficient expressiveness for the cipher-images. In this paper, we propose a novel paradigm named Encrypted Vision Transformer (EViT), which advances the discriminative representations capability of cipher-images. First, in order to capture comprehensive ruled information, we extract multi-level local length sequence and global Huffman-code frequency features from the cipher-images which are encrypted by stream cipher during JPEG compression process. Second, we design the Vision Transformer-based retrieval model to couple with the multi-level features, and propose two adaptive data augmentation methods to improve representation power of the retrieval model. Our proposal can be easily adapted to unsupervised and supervised settings via self-supervised contrastive learning manner. Extensive experiments reveal that EViT achieves both excellent encryption and retrieval performance, outperforming current schemes in terms of retrieval accuracy by large margins while protecting image privacy effectively. Code is publicly available at \url{https://github.com/onlinehuazai/EViT}.
\end{abstract}



\maketitle

\section{Introduction}

 Image retrieval is an important research field in the computer vision community, which draws increasing attention due to its significant research impacts and tremendous practical values \cite{abs-2101-11282, LiuLQWT16, 10.1145/3065950}. Given a query image, the objective of image retrieval is to search similar images in an extensive image database. With the rise of cloud computing, traditional local storage mode has been changed to cloud storage, which satisfies user demand for storing massive image data on the server and enables users to access the data from any location at any time. Although cloud computing contributes to alleviating the challenge of limited local storage space and provides great convenience to users, the images are at danger of privacy leakage since the cloud server cannot be fully trusted and is vulnerable to hacking \cite{XiaWTXW21}. A typical strategy is to encrypt the images prior to uploading them to the cloud server, but conventional image encryption algorithms may hinder the subsequent encrypted image data retrieval operation \cite{LiS19, zhang2014histogram}. Therefore, it is urgent to develop privacy-preserving image retrieval technology that can provide both privacy protection and accurate retrieval simultaneously.
	
    \begin{table}[hb]
		\centering
		\captionsetup{font=footnotesize}
		\caption{Our scheme and the current schemes.}
		\label{tab1}
        \begin{threeparttable}
			\begin{tabular}{@{}lcccccc@{}}
				\toprule
				\multirow{3}{*}{Schemes} &
				\multicolumn{3}{c}{Image encryption \& Feature extraction}   & \multicolumn{3}{c}{Retrieval model}  \\ 
				\cmidrule(l){2-7} &
				Visual & Adaptive & Multi-level & 
				\multicolumn{1}{c}{\multirow{2}{*}{Unsupervised}} & 
				\multicolumn{1}{c}{\multirow{2}{*}{Supervised}} & 
				\multicolumn{1}{c}{\multirow{2}{*}{Deep learning}} \\
				~ & security & encryption key & features \\
				\midrule
				Zhang \cite{ zhang2014histogram} & --                      & \checkmark  & \ding{56} & \checkmark  & \ding{56} & \ding{56} \\ \midrule
				Liang \cite{ liang2019huffman}   & $\uparrow$ & \ding{56} & \ding{56} & \checkmark  & \ding{56} & \ding{56} \\ \midrule
				Xia \cite{xia2019privacy}        & $\uparrow$ & \ding{56} & \ding{56} & \checkmark  & \ding{56} & \ding{56} \\ \midrule
				Li \cite{LiS19}                  & $\uparrow$ & \checkmark  & \ding{56} & \checkmark  & \ding{56} & \ding{56} \\ \midrule
				Xia \cite{XiaWTXW21}             & $\uparrow$ & \ding{56} & \ding{56} & \checkmark  & \ding{56} & \ding{56} \\ \midrule
				Xia \cite{XiaJLLJ22}             & $\uparrow$ & \ding{56} & \ding{56} & \checkmark  & \ding{56} & \ding{56} \\ \midrule
				Chen \cite{ cheng2016markov}     & $\uparrow$ & \checkmark  & \ding{56} & \ding{56} & \checkmark  & \ding{56} \\ \midrule
				Feng \cite{FengLLLH21}           & $\uparrow$ & \ding{56} & \ding{56} & \ding{56} & \checkmark  & \checkmark  \\ \midrule
				Ours                             & $\uparrow$ & \checkmark  & \checkmark  & \checkmark  & \checkmark  & \checkmark  \\ \bottomrule
		\end{tabular}
      \begin{tablenotes}
       \footnotesize
       \item[1] ``\checkmark'' indicates that corresponding requirements are supported; otherwise, `\ding{56}'' is used.
       \item[2] ``$\uparrow$'' indicates better visual security than Zhang \cite{zhang2014histogram} (mentioned in Section \ref{experiments}).
     \end{tablenotes}
  \end{threeparttable}
	\end{table}

	Existing privacy-preserving image retrieval schemes can be roughly classified into two categories: feature-encryption-based and image-encryption-based schemes \cite{XiaWTXW21}. In the first category, the content owner first extracts features (e.g.,  scale-invariant feature transform) from plain-images, then separately encrypts images and features \cite{LuVSW09, XiaWZQSR16, XiaXVS17, janani2021secure}. In order to match cipher-images, the encrypted features of similar images should keep distance similarity \cite{XiaJGYX22}, and both encrypted images and features are uploaded to the server. This type of schemes can effectively protect privacy with standard cryptographic technologies, however, it performs image encryption and feature extraction/encryption separately, which may incur additional computational workload and inconvenience for the owner and users \cite{XiaWTXW21, LiS19, zhang2014histogram}. To address it, image-encryption-based schemes were proposed by extracting features from encrypted images directly \cite{XiaWTXW21, XiaJLLJ22, LiS19, xia2019privacy, liang2019huffman, cheng2016ac, zhang2014histogram, cheng2016encrypted, cheng2016markov}. In such schemes, the content owner only needs to encrypt images, and then uploads encrypted images to the server. The task of extracting features from cipher-images can be outsourced to the server, which can decrease computational workload for owner and users. There are mainly three modules in the system of the second type of schemes, namely image encryption algorithm, feature extraction method and retrieval model. The mentioned three modules are inherently correlated to each other. The image encryption algorithm is the foundation, which retains effective ruled features in cipher-images and provides desirable image security. Feature extraction method is the bridge, which offers comprehensive inputs for retrieval model and determines whether adaptive encryption key \cite{HeHTH18} is supported or not (the same image can be encrypted with different secret keys and the extracted features are unchanged before and after encryption). The focus of retrieval model is to achieve excellent retrieval accuracy, which needs to couple with the extracted features and learn discriminative  representations for the cipher-images. 

    Our work follows the idea of image-encryption-based privacy-preserving image retrieval, which can reduce additional computational workload. Zhang \textit{et al.} \cite{zhang2014histogram} proposed the first image-encryption-based privacy-preserving scheme, while they had poor visual security with simple permutation encryption \cite{liang2019huffman}. In order to achieve better visual security, state-of-the-art schemes \cite{liang2019huffman, xia2019privacy, XiaWTXW21, XiaJLLJ22, LiS19, cheng2016markov, FengLLLH21} introduced sophisticated encryption operations (e.g., value replacement and stream cipher) to encrypt images, but still have some deficiencies in feature extraction methods and retrieval models. For example, \cite{zhang2014histogram, liang2019huffman, xia2019privacy, XiaWTXW21, XiaJLLJ22, LiS19, cheng2016markov} just extract shallow features (e.g., histogram features) from cipher-images, which are unable to express enough information. Furthermore, \cite{liang2019huffman, xia2019privacy, XiaWTXW21, XiaJLLJ22} fail to support the adaptive encryption key \cite{HeHTH18} because the feature spaces are randomly changed with different secret keys. Feng \textit{et al.} \cite{FengLLLH21} divided images into $8 \times 8$ blocks and proposed Vision Transformer \cite{DosovitskiyB0WZ21} (ViT) based supervised retrieval model, which achieves better retrieval performance than convolutional neural network (CNN), but it is painstaking to assign labels to images. Schemes \cite{liang2019huffman, xia2019privacy, XiaWTXW21, XiaJLLJ22, zhang2014histogram, LiS19} utilized trivial models (e.g., K-means \cite{macqueen1967some} and Bag-of-Words (BOW) \cite{SivicZ03}) to build unsupervised retrieval models, but it is difficult for these trivial models to learn non-linear embedding of complex image datasets \cite{XieGF16}, while deep neural network (DNN) is skilled in it. To capture more plentiful information of cipher-images, our feature extraction method designs multi-level features from $8 \times 8$ blocks and global Huffman-code of cipher-images. We utilize the length of variable-length integer (VLI) code unchanged with stream cipher to support the adaptive encryption key. Self-supervised learning is the typical unsupervised framework \cite{ChenK0H20, He0WXG20, TianKI20,GansbekeVGPG20, DangD0WH21, JangC21}, and ViT \cite{DosovitskiyB0WZ21} divides an image into non-overlapping blocks to learn global dependency relations with self-attention mechanism \cite{VaswaniSPUJGKP17}. Hence, inspired by Feng \cite{FengLLLH21}, and ViT can couple with our multi-level features from $8 \times 8$ blocks of cipher-images, we build ViT-based unsupervised retrieval model in self-supervised learning manner.

	In this paper, we propose a novel privacy-preserving image retrieval scheme named Encrypted Vision Transformer (EViT). First, images are encrypted during JPEG compression process, in which the VLI code of Discrete Cosine Transform (DCT) coefficient is encrypted by stream cipher. Second, EViT extracts well-designed multi-level features from cipher-images: the length sequence of DCT coefficients’ VLI code in each 8 $\times$ 8 block (local features) and the global Huffman-code frequency features, which can express more plentiful information of cipher-images. The VLI code encryption with stream cipher has an inherent advantage: the length of VLI code before and after encryption are unchanged, so multi-level features remain invariable whichever secret keys are used, namely EViT supports the adaptive encryption key. Finally, EViT adopts self-supervised contrastive learning \cite{ChenK0H20, He0WXG20, abs-2003-04297} manner to build the unsupervised retrieval model, which can reduce label overhead. In order to learn discriminative representations of the cipher-images, EViT proposes modified Vision Transformer-based retrieval model, and replaces original $Cls\_Token$ \cite{DosovitskiyB0WZ21} of ViT with learnable global Huffman-code frequency feature. Conventional data augmentations in the plain-image retrieval domain (e.g., random crop) will entirely change the multi-level features of cipher-images (Section \ref{data aug}), so EViT directly takes two adaptive data augmentations for multi-level features to improve representation ability of retrieval model. EViT can also achieve the supervised retrieval model by easily Fine-Tuning on the unsupervised model. 
 
    As shown in Tab. \ref{tab1}, EViT can satisfy all requirements (e.g., adaptive encryption key, multi-level features, deep learning) simultaneously, and our experimental results show that our EViT can improve retrieval performance by large margins. The main contributions are summarized as follows:

	\begin{itemize}
		\item[1)]
		Ingenious multi-level features, local length sequence and global Huffman-code frequency, are extracted from cipher-images to express more abundant features of cipher-images. Our feature extraction method also enables EViT to use the adaptive encryption key since the length of VLI code is unaffected by stream cipher.
	\end{itemize}

	\begin{itemize}
		\item[2)]
		To the best of our knowledge, EViT is the first to propose self-supervised learning for privacy-preserving image retrieval.  In order to enhance model's representation ability, EViT proposes two adaptive data augmentations for retrieval model, and uses learnable global Huffman-Code frequency to improve existing ViT.
	\end{itemize}
	
	\begin{itemize}
		\item[3)]
		Experimental results demonstrate that our EViT outperforms other state-of-the-art schemes in retrieval performance for both unsupervised and supervised learning models. EViT can not only protect image privacy but also complete image retrieval efficiently. 
	\end{itemize}
	
	The rest of our paper is organized as follows. Section \ref{related work} presents the related work for privacy-preserving image retrieval. Preliminaries are introduced in Section \ref{preliminaries}. Section \ref{proposed scheme} gives our proposed scheme. Section \ref{experiments} presents the experimental results. Finally, Section \ref{conclusion} summaries the conclusion.

\section{Related work}
	\label{related work}
	
	In recent years, researchers have paid more attention on privacy-preserving image retrieval \cite{LuVSW09, LuSVW09, XiaWZQSR16, XuGXXWS17, XiaJLLJ22, LiuSXS17}, and applied these techniques to boost the performance of real-life applications \cite{weng2014privacy, OsadchyPJM10, weng2016privacy}. The current privacy-preserving image retrieval schemes mainly can be divided into two categories \cite{XiaWTXW21}: feature-encryption-based and image-encryption-based schemes.
	
	\textit{\textbf{Feature-encryption-based schemes}}: In this type of works, content owner first extracts features from plain-images, then encrypts these features and images separately. It's noted that the features of plain-images also need to be protected because the features always contain sensitive information of plain-images \cite{XiaJGYX22, BenhamoudaHH19}. Lu \textit{et al.} \cite{LuVSW09} proposed the first privacy-preserving image retrieval scheme, and they improved retrieval speed based on safe search indexes in \cite{LuSVW09}. Wai \textit{et al.} \cite{WongCKM09} developed a new asymmetric scalar-product-preserving encryption (ASPE) that preserved a special type of scalar product which could provide k-nearest neighbor (kNN) computation on encrypted features. Xia \textit{et al.} \cite{XiaWZQSR16} utilized a stable KNN algorithm to protect image feature vectors and designed a watermark-based protocol to prevent the authorized query users from illegally distributing the retrieved images. Zhang \textit{et al.} \cite{ZhangJLLDGL17} used fully homomorphic encryption to encrypt the features and proposed to boost privacy-preserving search by distributed and parallel computation. Xia \textit{et al.} \cite{XiaZSQR18} proposed a scheme with extracting features by the Scale-invariant feature transform (SIFT) feature and  BOW model. They calculated the distance among image features through the Earth Mover's Distance (EMD) and adopted a linear transformation on EMD to protect parameter information. Cheng \textit{et al.} \cite{ZhangZZY20} used hash codes to encrypt features which were learned by deep neural networks (DNN), and they utilized S-Tree to increase the search efficiency. Feature-encryption-based schemes can achieve great security of images, but it performs image encryption and feature extraction/encryption separately, resulting in an additional computational workload and inconvenience for the content owner and users \cite{LiS19, liang2019huffman}.

	\textit{\textbf{Image-encryption-based schemes}}: This type of work extracts features directly from cipher-images. Content owner only needs to encrypt images before uploading cipher-images to the server, the task of extracting features from cipher-images can be outsourced to the server, which solves the problem of separation of image encryption and feature extraction/encryption. Zhang \textit{et al.} \cite{zhang2014histogram} proposed a scheme with encryption of JPEG images by permuting DCT coefficients and extracting features from these coefficients’ histogram invariance. Cheng \textit{et al.} \cite{cheng2016ac} proposed to encrypt the DC coefficients by using stream cipher, encrypt the AC coefficients by using scrambling encryption, and conduct retrieval based on the histogram of the AC coefficients. However, Cheng \textit{et al.} \cite{cheng2016ac} could not ensure JPEG format compliance. Xia \textit{et al.} \cite{xia2019privacy} encrypted DC coefficients by stream cipher on the Y component and encrypted U and V components by value replacement and permutation encryption, then extracted AC histograms features of Y component and color histograms features of U and V components. Liang \textit{et al.} \cite{liang2019huffman} extracted Huffman-code histograms from cipher-images which were encrypted by stream cipher and permutation encryption. The work \cite{XiaJLLJ22} encrypted images by color value substitution and permutation encryption, and extracted local color histogram features from cipher-images, then they built unsupervised bag-of-encrypted-words (BOEW) model to achieve retrieval. Li \textit{et al.} \cite{LiS19} proposed a new block transform encryption method using orthogonal transforms rather than $8 \times 8$ DCT. \cite{XiaWTXW21} extracted secure Local Binary Pattern (LBP) features from cipher-images, then built BOW model to conduct retrieval. These above schemes built unsupervised retrieval model, but they fail to use deep learning to learn non-linear embedding of cipher-images, and just extracted shallow features from cipher-images. Cheng \textit{et al.} \cite{cheng2016markov} used stream cipher and permutation encryption to encrypt JPEG images, and extracted features from cipher-images by Markov process, then built supervised support vector machine (SVM) model to conduct retrieval. But it is painstaking to assign labels to images, and SVM is a linear model which is limited to learn discriminative representations of cipher-images. 
 
	Our EViT follows the idea of image-encryption-based privacy-preserving image retrieval which can decrease additional computational workload. EViT extracts multi-level features from cipher-images and introduces DNN based unsupervised retrieval model in a self-supervised manner, which can significantly improve retrieval performance.

\section{Preliminaries}
	\label{preliminaries}
	
	\subsection{JPEG Compression}
	
\begin{figure}[]
		\centering
		\includegraphics[width=0.75 \textwidth]{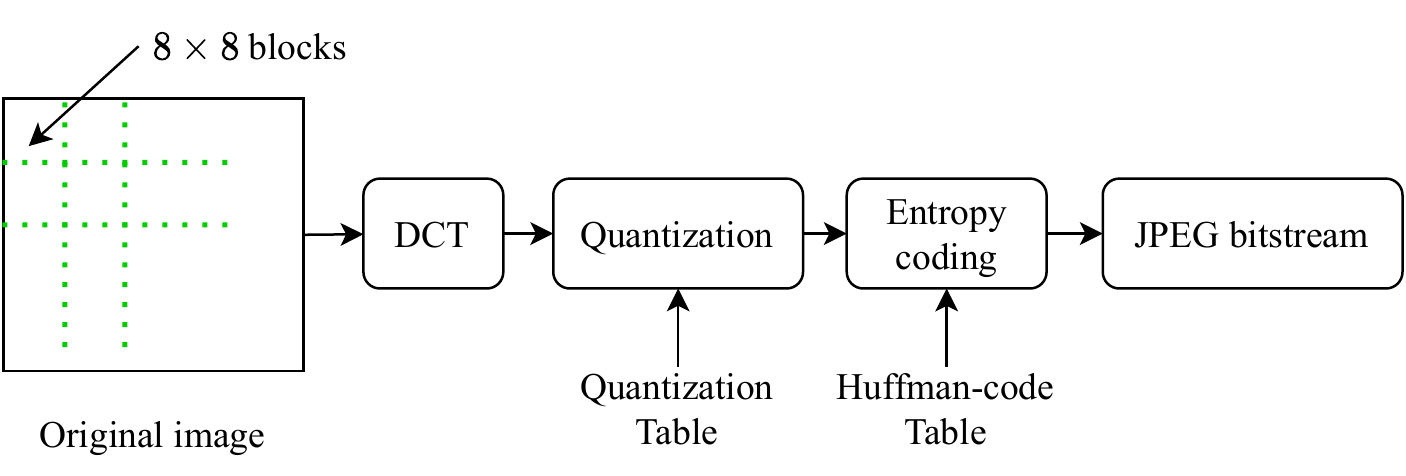}
		\captionsetup{font=footnotesize}
		\caption{Overview of JPEG compression process.}
		\label{JPEG_compression}
	\end{figure}

	We encrypt images during JPEG compression process like some privacy-preserving image retrieval schemes \cite{zhang2014histogram, cheng2016ac, xia2019privacy, liang2019huffman, LiS19, XiaJGYX22}. Here, we briefly introduce the JPEG compression process, whose overview is shown in Fig. \ref{JPEG_compression}. According to the JPEG compression standard \cite{pennebaker1992jpeg, ChristopoulosES00}, images are converted to YUV color space. After $8 \times 8$ DCT and quantization, each block has a total of 64 coefficients, of which the first coefficient is the direct current (DC) coefficient, and the remaining 63 coefficients are the alternating current (AC) coefficients. The DC coefficient is encoded by differential pulse code modulation (DPCM), and the remaining 63 AC coefficients in the same block are converted into a sequence using zig-zag scanning. AC coefficients are encoded with run-length encoding (RLE), which are converted to the $(r,v)$ pairs \cite{pennebaker1992jpeg, ChristopoulosES00}. 
	
\begin{figure}[]
		\centering
		\includegraphics[width=0.75 \textwidth]{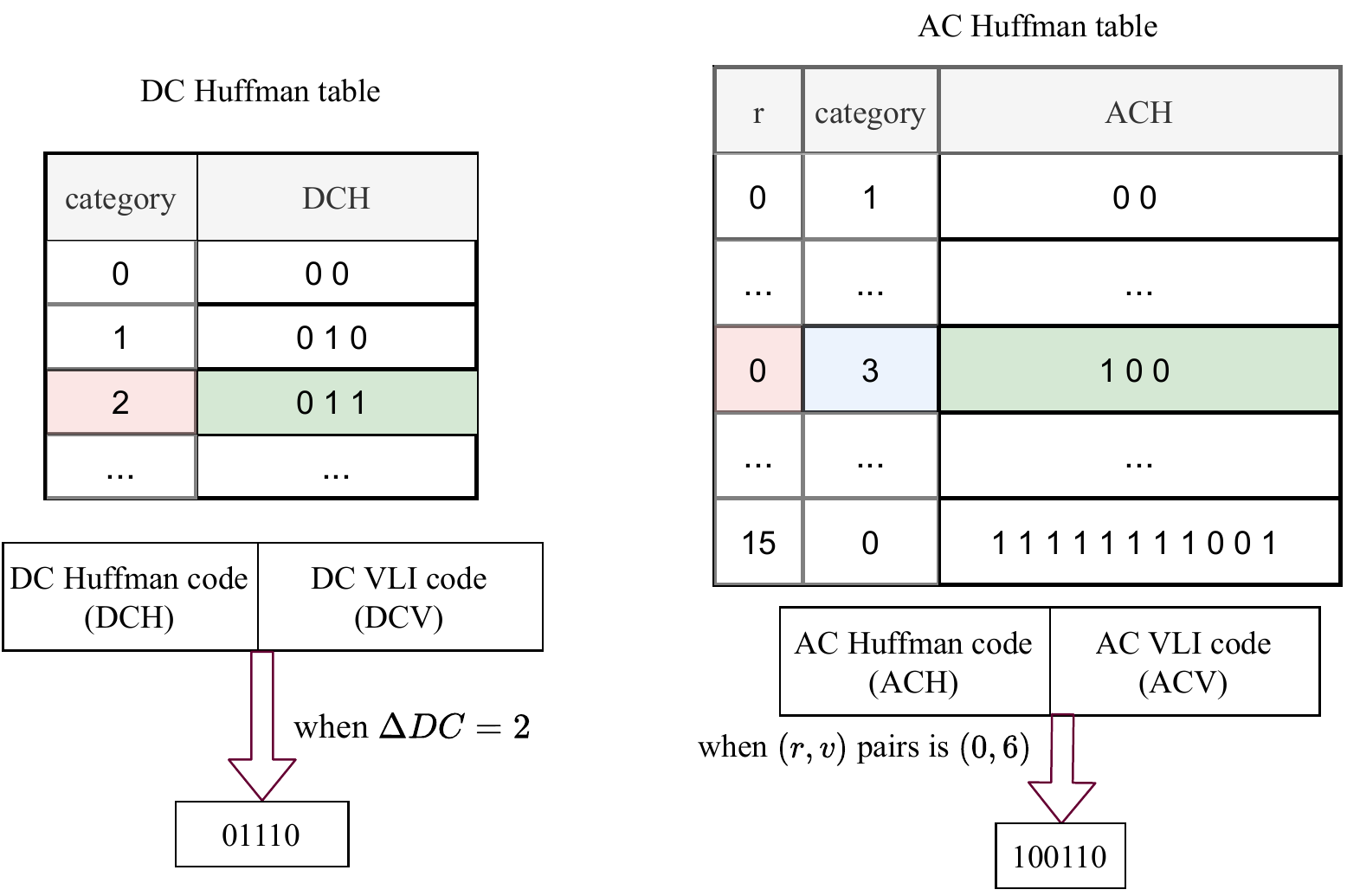}
		\captionsetup{font=footnotesize}
		\caption{An example of lossless Huffman variable-length entropy coding.}
		\label{huffmancode}
	\end{figure}

	The lossless Huffman variable-length entropy coding technology is used to further compress the DC differential ($\Delta DC$) and AC coefficients $(r,v)$ pairs, and all coefficients are coded to binary sequence. Specifically, each $\Delta DC$ is encoded into two parts: DC Huffman code (DCH) and DC VLI code (DCV); each $(r,v)$ pairs is encoded as two parts: AC Huffman code (ACH) and AC VLI code (ACV). As shown in Fig. \ref{huffmancode}, we take an example for lossless Huffman variable-length entropy coding, where category is the range of amplitudes of coefficients \cite{pennebaker1992jpeg}, and $r$ corresponds to the run value in the AC Human table \cite{pennebaker1992jpeg}. For example, when $\Delta DC=2$, it's $DCH=011$ and $DCV=10$, so it is coded into $01110$; when $(r,v)=(0,6)$, it's $ACH=100$ and $ACV=110$, therefore it is coded into $100110$. Each coefficient can be coded into binary sequence through Huffman-code table mapping, more details please refer to \cite{pennebaker1992jpeg, ChristopoulosES00}.
	
	\subsection{Unsupervised Contrastive Learning}
	
	Supervised learning has been widely used in many image retrieval tasks \cite{LiuYVCRMG19,DengYLT20,HusainB19,YangLSWRY18,LvZTSL18,10.1145/3404835.3462812}, but it is painstaking to assign labels to images. In order to decrease the overhead with human annotations, unsupervised algorithms are explored by researchers, such as K-means \cite{macqueen1967some} and BOW \cite{SivicZ03} model. Both K-means and BOW utilize the idea of clustering, but in many image databases, it is ineffective to learn images' representations because clustering with euclidean distance cannot learn non-linear embedding \cite{XieGF16}. Deep neural networks (DNN) is competent in learning non-linear embedding which are indispensable for complex image databases. Supervised-learning based on DNN generally uses target labels to train model, but it is painstaking to assign labels to images. Self-supervised learning can build DNN model without target labels, which generates virtually unlimited labels from existing images and uses those to learn the representations. In recent years, many self-supervised learning methods have been proposed \cite{GidarisSK18, RenL18, NorooziF16, BojanowskiJ17, WuXYL18, He0WXG20, ChenK0H20, TianKI20, MisraM20}. Due to its simple and effective property, self-supervised contrastive learning \cite{WuXYL18, ChenK0H20, He0WXG20, TianKI20} has been used in many  unsupervised-learning computer vision tasks \cite{GansbekeVGPG20, DangD0WH21, JangC21, abs-2112-01390, abs-2105-04553, abs-2104-03602}. SimCLR \cite{ChenK0H20} is a popular self-supervised contrastive learning method, whose process is roughly illustrated as follows: given an image $x$, we can obtain different images $x_{i}$ and $x_{j}$ by stochastic data augmentations such as random color distortions and random Gaussian blur; then the two images through same encoder can generate corresponding embeddings $z_{i}$ and $z_{j}$, and SimCLR can learn representations of images by maximizing the similarity $z_{i}$ and $z_{j}$. SimCLR obtains the positive samples of $x$ by data augmentations, the other images are negative samples. But SimCLR \cite{ChenK0H20} needs a very large batch size to build negative samples when training, this is expensive for most researchers to use GPU with large memory. Therefore MoCo \cite{He0WXG20, abs-2003-04297} proposed momentum contrast to solve the problem of large batch size by building a dynamic dictionary with a queue and momentum updating. In this paper, we utilize self-supervised contrastive learning based on MoCo to build our unsupervised-learning retrieval model.
	
	\subsection{Vision Transformer}
    \label{ViT}
	
	Since the proposal of Google's Transformer \cite{VaswaniSPUJGKP17} in 2017, it has almost dominated natural language processing tasks \cite{DevlinCLT19, RaffelSRLNMZLL20, FarahaniGFM21}. Transformer is a sequence model based on self-attention mechanism, which can capture the global dependencies between words. In 2021, Google proposed to apply Transformer in computer vision \cite{DosovitskiyB0WZ21}, called Vision Transformer (ViT). Then many researchers explore ViT and find ViT can obtain excellent performance in many computer vision tasks \cite{LinW021, KimLKKK21, PrakashC021, 0012WZXXT21, SunSWBZ21, 10.1145/3534678.3539322}, even surpass the CNN model in performance \cite{LiuL00W0LG21, abs-2105-04553, GrahamETSJJD21, WangX0FSLL0021, Arnab0H0LS21}. Vision Transformer not only makes breakthroughs in computer vision but also forms the unified model for computer vision and natural language processing. 
	
	ViT \cite{DosovitskiyB0WZ21} divides images into non-overlapping blocks, and each block is equal to a word in natural language processing. Suppose image's size is $H \times W$, the size of each block is $P \times P$, hence the number of blocks is $\frac{H\times W}{P^{2}}$. Assuming that the dimension of word embedding is $D$, ViT uses linear projection \cite{DosovitskiyB0WZ21} to map each block to dimension $D$, called block embedding. Like the $[class]token$ in BERT \cite{DevlinCLT19}, ViT prepends a learnable embedding $Cls\_Token$ in block embeddings. The goal of $Cls\_Token$ is to learn the representation of image, and ViT initializes $Cls\_Token$ with ones (dimension $D$). The results of adding the block and position embeddings \cite{DosovitskiyB0WZ21, VaswaniSPUJGKP17} are then used as the input to the Transformer encoder. The Transformer encoder of ViT is similar with standard Transformer \cite{VaswaniSPUJGKP17}, includes self-attention \cite{VaswaniSPUJGKP17}, layer normalization \cite{BaKH16} and residual module \cite{HeZRS16}. Self-attention is an import part in the Transformer encoder. When calculating self-attention, three matrices are needed: query (Q), key (K), value (V). Suppose the input of self-attention is $X$, and the definition of self-attention is expressed as:
	\begin{equation}
		Q=W_{Q}X, \quad K=W_{K}X, \quad V=W_{V}X
	\end{equation}
	\begin{equation}
		Attention(Q,K,V)=softmax(\frac{QK^{T}}{\sqrt{d_{k}}})V
	\end{equation}
	where $W_{Q}, W_{K}, W_{V}$ are linear projection matrices, and $d_{k}$ is the dimension of $K$. In order to capture more information, Transformer uses multi-head self-attention to learn different subspaces \cite{VaswaniSPUJGKP17}. Multi-head self-attention uses $h$ different linear projections to map $Q, K, V$, and then concatenates different results of self-attention and does a linear projection. The definition of multi-head self-attention is as follows:
	\begin{equation}
	MultiHead(Q,K,V)=Concat(head_{1},\dots,head_{h})W^{O}
			\label{MSA}
	\end{equation}
    \begin{equation}
		head_{i}=Attention(Q_{i},K_{i},V_{i}), \quad i\in[1,h]
	\end{equation}
	where $W^{O}$ is linear projection matrices. 
	
	In this paper, we use ViT as backbone in our retrieval model, and the reasons are as follows: a) ViT is popular and makes excellent performance in recent many computer vision tasks \cite{LiuL00W0LG21, abs-2105-04553, GrahamETSJJD21, WangX0FSLL0021, Arnab0H0LS21, LinW021, KimLKKK21, PrakashC021, 0012WZXXT21, SunSWBZ21}. b) Our feature extraction method couples with the inputs of ViT. We extract features from cipher-images' $8 \times 8$ blocks, and ViT divides image into non-overlapping blocks. So the features of each $8 \times 8$ block are equal to word embedding of ViT. c) ViT can learn global dependency relations with self-attention mechanism \cite{VaswaniSPUJGKP17} compared with local CNN \cite{abs-2105-10497}. Previous privacy-preserving image retrieval scheme \cite{FengLLLH21} also used ViT as backbone and proved that ViT was more fit to encrypted images than CNN in their experiments.

	\begin{figure}[hb]
		\centering
		\includegraphics[width=0.5 \textwidth]{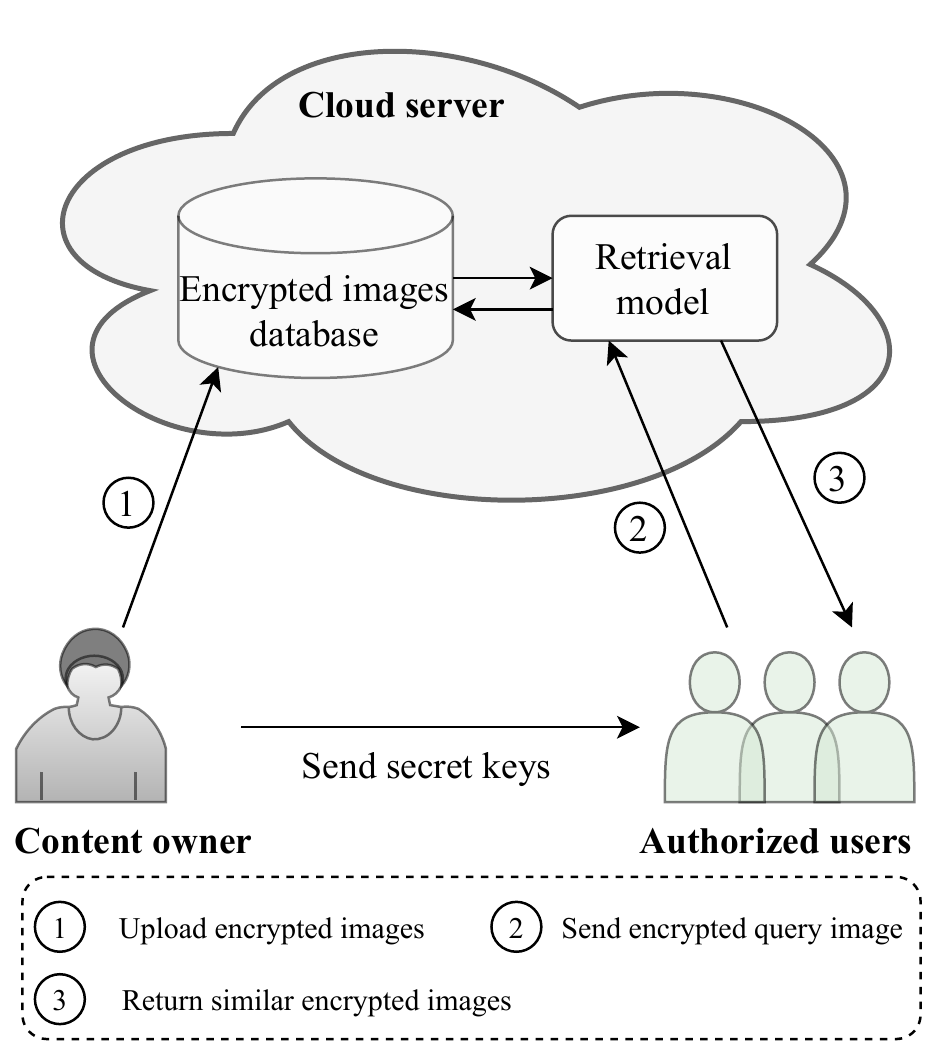}
		\captionsetup{font=footnotesize}
		\caption{System model of image-encryption-based privacy-preserving image retrieval.}
		\label{fig_1}
	\end{figure}

\section{Proposed Scheme}
	\label{proposed scheme}
     Generally, the system of image-encryption-based privacy-preserving image retrieval includes three parts: content owner, server, and authorized users. As shown in Fig. \ref{fig_1}, the content owner first encrypts images and uploads encrypted images to the server. Then authorized users encrypt the query image with the same encryption algorithm and upload it to the server. The server extracts features from the query image, and the retrieval model searches similar cipher-images in encrypted image database according to the features. Finally, these similar cipher-images are returned to authorized users, and authorized users decrypt images with encryption keys which are shared from the content owner through a secure channel \cite{liang2019huffman,  cheng2016markov}. In this system, three modules need to be designed: image encryption algorithm, feature extraction method, and image retrieval model. We specify these modules of EViT in detail below.
 
	\subsection{Image Encryption}
	\label{proposed_encryption}
	
	Images are encrypted during JPEG compression process which is explained briefly in Section \ref{preliminaries}. Specifically, in the stage of entropy coding, we take stream exclusive-or operator for VLI code of DC and AC coefficients \cite{QianZW14}. The corresponding encryption keys are $k_{DC}$ and $k_{AC}$, and we encrypt DCV and ACV as follows:
	\begin{equation}
		DCV^{'}\leftarrow DCV \oplus k_{DC}
	\end{equation}
	\begin{equation}
		ACV^{'}\leftarrow ACV \oplus k_{AC}
	\end{equation}
	where $DCV^{'}$ and $ACV^{'}$ are encrypted VLI codes, $\oplus$ is exclusive-or operator. Our encryption keys are generated by the adaptive encryption key generation method \cite{HeHTH18} which uses image as input of hash function BLAKE2 \cite{AumassonNWW13}, therefore different images have different encryption keys. During encryption process, original image $I$ can be compressed to encrypted JPEG bitstream, and different color spaces of $I$ have different secret keys. Stream exclusive-or with VLI code theoretically does not increase storage memory of cipher-images \cite{ChristopoulosES00}. In Algorithm \ref{algorithm_encryption}, the encryption algorithm of our proposed EVIT is presented. The encrypted bitstream is decodable because the file structure and Huffman-code are unchanged \cite{QianZW14}.
	
	\IncMargin{1em}
	\begin{algorithm}[]
		\caption{Encryption algorithm}
		\label{algorithm_encryption}
		\small
		\SetKwInOut{Input}{input}\SetKwInOut{Output}{output}
		\Input {$I$, $k_{DC}$, $k_{AC}$} 
		\Output {Encrypted JPEG bitstream}
		\BlankLine 
		Convert $I$ from RGB to YUV color space\;
        Denote the width and height of the image $I$ as $W$ and $H$ \;
		\For {$I_{i}$ in $I$, $i \in Y,U,V$}
		{
			Divide $I_{i}$ into several $8 \times 8$ non-overlapping blocks $B^{i}_{j}$, $j\in [1, \dots, blknum]$, where $blknum=\frac{W\times H}{8\times 8}$\; 
			Conduct DCT, quantization and entropy coding\;
			\For{$j=1$ to $blknum$}
			{
				Encrypt the VLI code of $B^{i}_{j}$\; 
				\begin{footnotesize}
					$ DCV^{'}_{B^{i}_{j}}  \leftarrow DCV_{B^{i}_{j}}  \oplus k_{DC}^{B^{i}_{j}}$\;
				\end{footnotesize}
				\begin{footnotesize}
					$ ACV^{'}_{B^{i}_{j}} \leftarrow ACV_{B^{i}_{j}}  \oplus k_{AC}^{B^{i}_{j}}$\;
				\end{footnotesize}
			}
		}
	\end{algorithm}
	\DecMargin{1em}

	\subsection{Feature Extraction}
	\label{feature_extraction}
	
	Image-encryption-based schemes extract features directly from cipher-images. Existing schemes just extract shallow features (e.g. DCT histogram), which are unable to express plentiful information of cipher-images. EViT extracts multi-level features from our cipher-images: local length sequence and global Huffman-code frequency features.
	
\begin{figure}[]
		\centering
		\includegraphics[width=0.75 \textwidth]{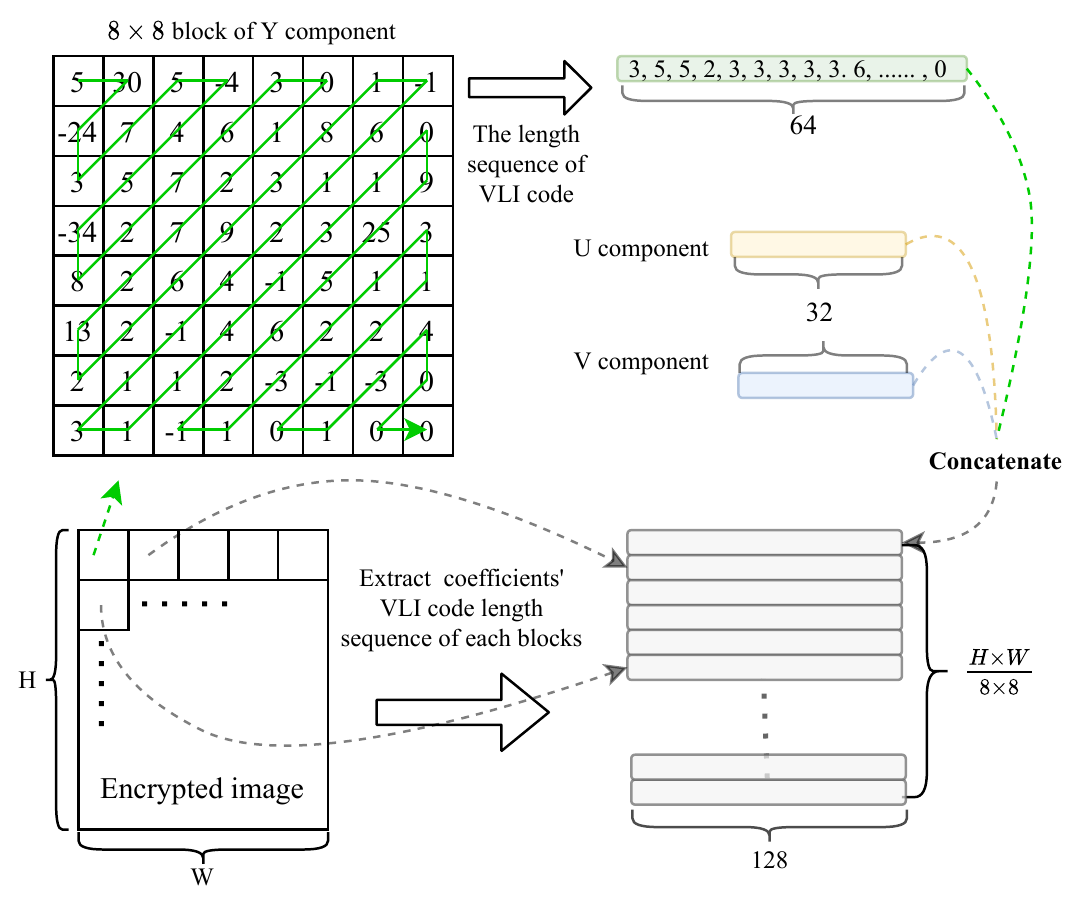}
		\captionsetup{font=footnotesize}
		\caption{Sketch of extracting local features from our encrypted images.}
		\label{extract_features}
	\end{figure}

	\textit{Local length sequence features:} EViT extracts each $8 \times 8$ block's local features. As shown in Fig. \ref{extract_features}, EViT calculates the corresponding VLI code's length of DCT coefficients and builds length sequence features in each $8 \times 8$ block. For example, when $\Delta DC=5$, it's VLI code is `101', hence the length is 3. The length sequence features are generated by zig-zag scanning \cite{pennebaker1992jpeg}. If the coefficient is zero, then EViT denotes its length as zero. Because each block has three components (Y/U/V), EViT concatenates three length sequence features of different components. It's noted that when extracting length sequence features for U and V components, EViT just chooses top $32$ coefficients in length sequence features due to there are many zeros of U and V components in the back coefficients \cite{ChristopoulosES00}.
	
	\textit{Global Huffman-code frequency features:} EViT extracts global Huffman-code frequency features from the cipher-images. We take a simple example to describe Huffman-code frequency features, as shown in Fig. \ref{huffmancode}. If the second row of DC Huffman table is used 10 times during entropy coding stage for an encrypted image, the corresponding Huffman-code frequency feature is denoted as 10. The rows of DC Huffman table is 12, and the rows of AC Huffman table is 162 \cite{pennebaker1992jpeg}. Therefore, the dimension of Huffman-code frequency features is  $(12+162)\times 3 = 522$, where $3$ represents three components (YUV).

	Global features and local features are extracted from cipher-images, which are employed to train our retrieval model. Here, different encryption keys can be utilized for different images since the stream exclusive-or operator does not change the length of VLI code. For example, suppose the DCV is `101', and the different encryption keys are `100' and `011', then the $DCV^{'}$ are `001' and `110' respectively. Hence the length of binary VLI code remains unchanged no matter which encryption key is used, namely the features that extracted from cipher-images are unchanged. This also means that we can use different encryption keys for different images, and the authorized user can encrypt query image with the same encryption algorithm but different keys.

	\subsection{Unsupervised-learning Retrieval Model}
	\label{module_3}
	After extracting features from encrypted images, EViT uses these features to train retrieval models. EViT respectively proposes the unsupervised-learning and supervised-learning retrieval models based on deep learning. Deep image retrieval model \cite{abs-2101-11282} is a typical deep metric learning \cite{KayaB19} whose aim is to learn the representations of images. Given an image, we first use learnable deep neural networks $f(\cdot)$ to learn it's representation $h$, and $f(\cdot)$ is generally called as backbone. Then the images' representations are used to calculate the similarity such as cosine or euclidean distances between the query and the targets. 
	
	\begin{figure}[]
		\centering
		\includegraphics[width=0.75 \textwidth]{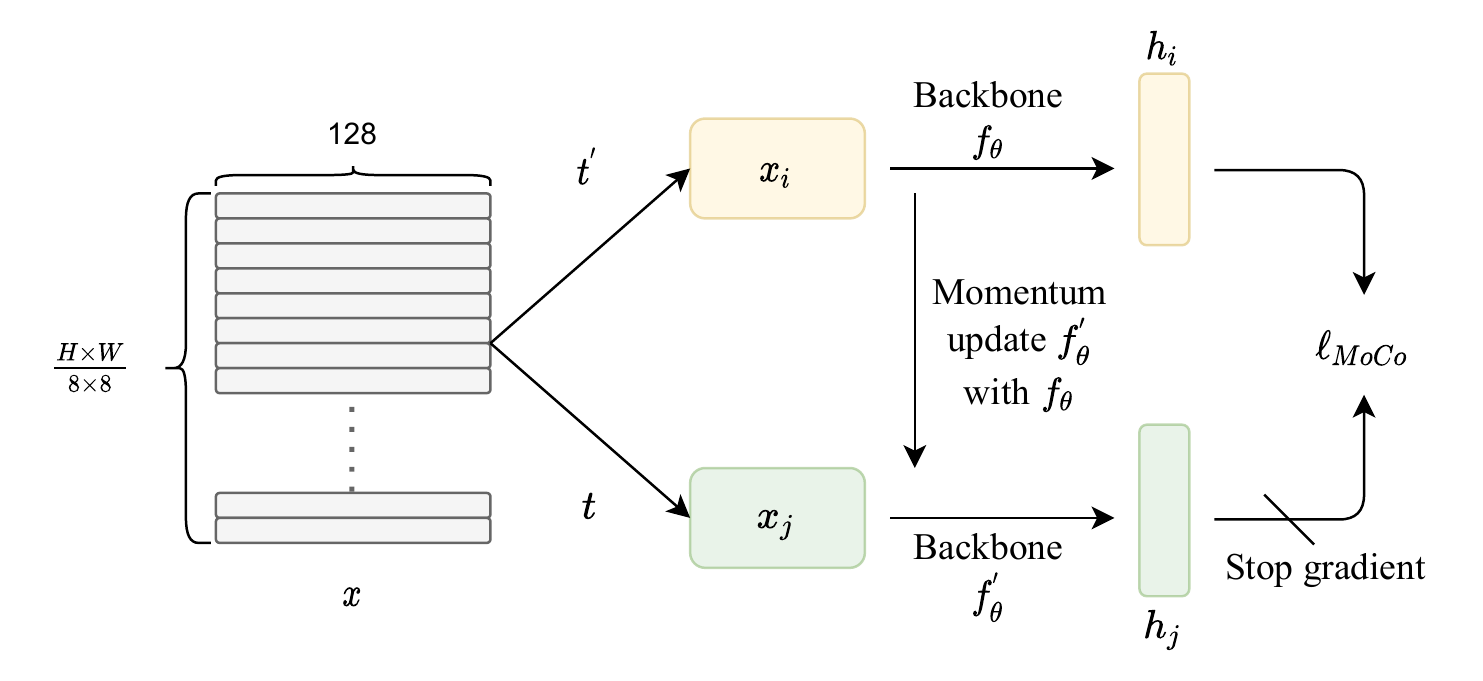}
		\captionsetup{font=footnotesize}
		\caption{Overview of our unsupervised-learning retrieval model.}
		\label{MoCov2}
	\end{figure}

	\subsubsection{Loss}
	
	The unsupervised framework uses MoCo (see Section \ref{preliminaries}) due to its simple and effective property \cite{ChenK0H20, He0WXG20, TianKI20}, which does not use target labels to learn the representations of cipher-images. As shown in Fig. \ref{MoCov2}, given a cipher-image, EViT extracts features from it and denotes these features as $x$. Using random data augmentations $t^{'}$ and $t$, we can obtain $x_{i}$ and $x_{j}$ respectively. Through backbone $f_{\theta}$, EViT can learn the representation $h_{i}$ of $x_{i}$. For $x_{j}$, the process is like $x_{i}$, but the backbone $f_{\theta}^{'}$ is stop-gradient which is updated by momentum with $f_{\theta}$ \cite{He0WXG20}. The process of forward propagation can be defined as:
	\begin{equation}
		x_{i}=t^{'}(x),\quad x_{j}=t(x)
	\end{equation}
	\begin{equation}
		h_{i} =f_{\theta}(x_{i}),\quad
		h_{j} =f_{\theta}^{'}(x_{j}).
	\end{equation}
	The backbone $f_{\theta}^{'}$ is updated with momentum manner \cite{He0WXG20} which is described as:
	\begin{equation}
		f_{\theta}^{'} = mf_{\theta}^{'}+(1-m)f_{\theta}
	\end{equation}
	where $m$ is momentum factor and is set to be $0.99$ following MoCo \cite{He0WXG20}. The structures of $f_{\theta}^{'}$ and $f_{\theta}$ are same, but with different parameters.

	The loss function is like MoCo which is called InfoNCE \cite{abs-1807-03748}. MoCo  proposed momentum contrast to solve the problem of large batch size by building a dynamic dictionary with a queue and momentum updating. The dynamic dictionary is a queue where the current batch enqueued and the oldest batch dequeued. For one sample $x_{i}$ in the current batch, the $x_{j}$ is positive sample, and other samples in the current batch and the queue are negative samples. The loss can be defined as:
	\begin{equation}
		\ell = -log\dfrac{exp(h_{i}\cdot h_{j}/\tau)}{exp(h_{i}\cdot h_{j}/\tau)+\sum_{k^{-}}exp(h_{i}\cdot h_{k^{-}}/\tau)}
	\end{equation}
	where $\tau$ is a temperature hyper-parameter proposed by \cite{WuXYL18}, which we set to be 0.1 like MoCo. $h_{i}$ and $h_{j}$ are representations of $x_{i}$ and $x_{j}$ respectively (Fig. \ref{MoCov2}). $h_{k^{-}}$ are representations of negative samples, and ``$\cdot$'' is dot product.

	\subsubsection{Backbone}
	\label{proposed_backbone}
	
	\begin{figure*}[]
		\centering
		\includegraphics[width=0.85 \textwidth]{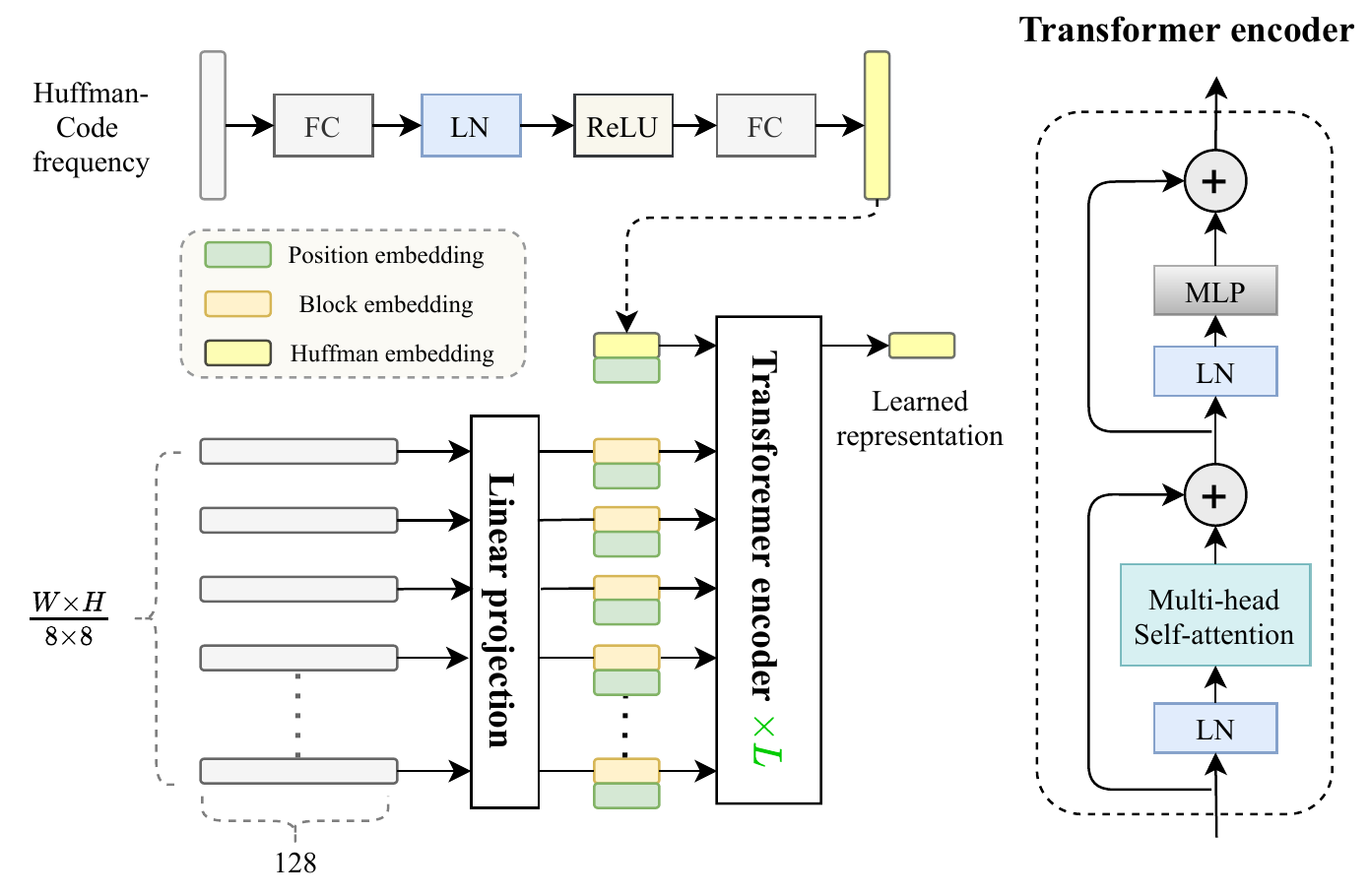}
		\captionsetup{font=footnotesize}
		\caption{Overview of our backbone.}
		\label{backbone}
	\end{figure*}
	
	Our backbone $f_{\theta}$ adopts the structure of ViT (see Section \ref{preliminaries}). EViT extracts two parts features from encrypted images, local length sequence features and global Huffman-code frequency features. As shown in Fig. \ref{backbone}, suppose the number of blocks of a cipher-image is $\frac{H\times W}{8 \times 8}$ ($H$ is height, $W$ is width). These local features, through linear projection \cite{DosovitskiyB0WZ21}, produce corresponding block embeddings. Original $Cls\_Token$ (mentioned in Section \ref{preliminaries}) of ViT are all ones for each image, which fails to express specific information for different images. Hence, different from standard ViT \cite{DosovitskiyB0WZ21}, EViT uses Huffman embedding to replace $Cls\_Token$, which is helpful for retrieval performance in experiments. The Huffman embedding ($He$) is learned from global Huffman-code frequency features ($gHff$), which can be defined as:
	\begin{equation}
		He=FC(\sigma(LN(FC(gHff))))
	\end{equation} 
	where FC is fully-connected layer, LN is layer normalization \cite{BaKH16}, $\sigma$ is activation function ReLU \cite{abs-1803-08375}. In order to keep position information, we also add position embedding \cite{DosovitskiyB0WZ21} with block embedding and Huffman embedding. The result of these embeddings is denoted as $v_{0}$, then through $L$ stacked Transformer encoder \cite{DosovitskiyB0WZ21}, EViT can learn the representations $v_{L}^{0}$ of cipher-image. The $l$-$th$ Transformer encoder can be defined as:
	\begin{equation}
		v_{l}^{'}=MSA(LN(v_{l-1}))+v_{l-1},\quad l=1,2,\dots L
	\end{equation} 
	\begin{equation}
		v_{l}=MLP(LN(v_{l}^{'}))+v_{l},\quad l=1,2,\dots L
	\end{equation} 
	where $MSA$ is multi-head self-attention \cite{VaswaniSPUJGKP17} (Eq. \ref{MSA}), MLP is multi-layer perceptron block \cite{DosovitskiyB0WZ21}. The output of $L$ stacked is $v_{L}$, and the representation is the first embedding $v_{L}^{0}$.
	
	ViT \cite{DosovitskiyB0WZ21} uses spatial pixels to build a block of plain-images, but spatial pixels are randomly encrypted in cipher-images. EViT uses length sequence feature to replace spatial pixels in a block. Different from plain-image retrieval which can use pre-trained ViT on imagenet \cite{KrizhevskySH12} as model's backbone, our task is privacy-preserving image retrieval and there is no pre-trained model as our backbone, so the retrieval model  needs to be trained from scratch. Generally, small learning rate and warm up \cite{HeZRS16} are necessary for training a new model with the structure of ViT \cite{TouvronCDMSJ21}. EViT uses cosine warm up \cite{WolfDSCDMCRLFDS20} with learning rate. Our experimental results also present that cosine learning rate with warm up is helpful for retrieval performance.

	\subsubsection{Data augmentation}
	\label{data aug}
	
	EViT uses random data augmentations $t$ and $t^{'}$ for encrypted features in our unsupervised-learning retrieval model (Fig. \ref{MoCov2}). Common plain-image data augmentations, such as random crop, are not suitable for our task. For example, the plain-image is a dog, we random crop an image from plain-image. The cropped image is equal to the plain image in which the spatial structure is the dog, and just the cropped image may not have the dog's tail. But in the encrypted images, the cropped image is not a dog after decryption. Because our encryption algorithm is conducted in JPEG compression process, each adjacent $8\times8$ block's DCT coefficients of encrypted images are correlational. If some blocks are missing, the DCT coefficients of encrypted cropped image are all changed during decryption process. Once DCT coefficients, in particular DC coefficients, are changed, the spatial pixels and structures of decrypted image will be changed correspondingly. Other plain-image data augmentations are also unsuitable for our task, because these augmentations all change DCT coefficients in augmented image. Thus EViT directly conducts data augmentations with our length sequence features extracted from cipher-images.

\begin{figure}[]
		\centering
		\includegraphics[width=0.75 \textwidth]{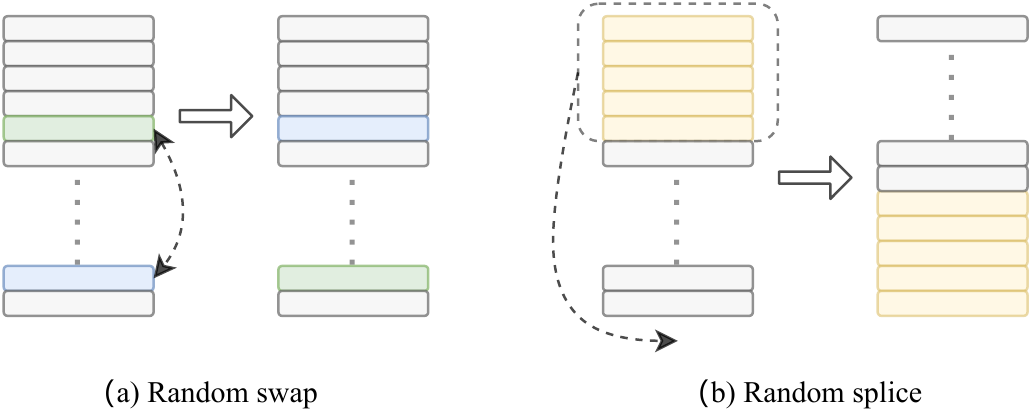}
		\captionsetup{font=footnotesize}
		\caption{Examples for our data augmentations with length sequence features.}
		\label{data_aug}
	\end{figure}

	In our retrieval model, we propose two kinds of adaptive data augmentations: random swap and splice for length sequence features. As shown in Fig. \ref{data_aug}, we take examples to describe these two kinds of data augmentations. Our data augmentations are motivated by two aspects: a) the length sequence features are like different words, swapping two words is a typical augmentation in nature language process \cite{WeiZ19}; b) ViT also can achieve well performance with simple block permutation \cite{abs-2105-10497}, and hence EViT can shuffle the length sequence features with random splice. In addition, model with random dropout \cite{SrivastavaHKSS14} also plays a role in data augmentation \cite{GaoYC21}. Because an image through model with random dropout twice can obtain different embeddings in the training stage \cite{GaoYC21}. ViT \cite{DosovitskiyB0WZ21} itself has a random dropout function, and EViT sets the dropout in our backbone as ViT \cite{DosovitskiyB0WZ21}. Our experimental results shows that the two adaptive data augmentations are vital for model, which can significantly improve retrieval performance.
	
	\subsection{Supervised-learning Retrieval Model}
	\label{supervised}
	
\begin{figure}[]
		\centering
		\includegraphics[width=0.75 \textwidth]{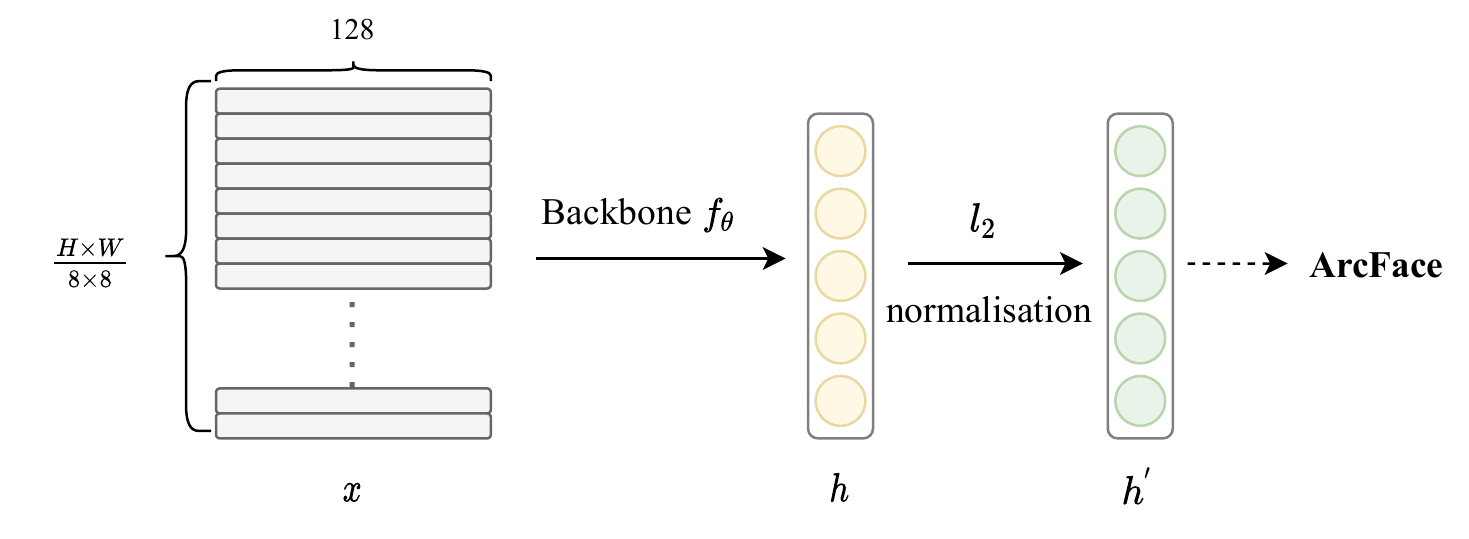}
		\captionsetup{font=footnotesize}
		\caption{Overview of supervised-learning retrieval model.}
		\label{supervised_model}
	\end{figure}

	Our EViT also provides a simple supervised-learning retrieval model, which uses the same structures of backbone like the unsupervised-learning model. As shown in Fig. \ref{supervised_model}, the supervised model can obtain representations $h$ of cipher-images after backbone $f_{\theta}$ (Fig. \ref{backbone}). Here we can use the pre-trained backbone from unsupervised-learning model. The supervised loss function ArcFace \cite{DengGXZ19} is used to train our model, which is defined as:
	\begin{equation}
		h=f_{\theta}(x), \quad h^{'}=l_{2}(h)
	\end{equation}
	where $l_{2}$ is short for $l_{2}$ normalisation \cite{DengGXZ19, WangXCY17}.
	
	ArcFace \cite{DengGXZ19} is a common deep metric learning loss function, which has been widely used in retrieval tasks \cite{abs-1906-04087, abs-2010-05350, DengZ19}. Compared with Triplet loss \cite{SchroffKP15}, ArcFace is more easy and effective \cite{DengGXZ19} which gets rid of the disadvantages such as hard sample mining and combinatorial explosion in the number of triplets. ArcFace adds an additive angular margin within softmax loss \cite{DengGXZ19}, which can be defined as: \par
	\begin{footnotesize}
		\begin{equation}
			\label{eq:arcface}
			\ell_{ArcFace}=-\frac{1}{N}\sum_{i=1}^{N}\log\frac{e^{s(\cos(\theta_{y_i}+\alpha))}}{e^{s(\cos(\theta_{y_i}+\alpha))}+\sum_{j=1,j\neq  y_i}^{n}e^{s\cos\theta_{j}}}
		\end{equation}
	\end{footnotesize}%
	where $N$ and $n$ are the batch size and the class number, and $\theta_{j}$ is the angle between the representation of $i$-th sample and $j$-th class center. $y_i$ is ground-truth of $i$-th sample, and $\theta_{y_i}$ represents the angle between $i$-th sample and ground-truth class center. $s$ and $\alpha$ are hyper-parameters which represent feature re-scale and angular margin parameters, respectively. More details please refer to \cite{DengGXZ19}.
	
	In the inference stage, we just need to calculate cosine distances of the representations $h$ of cipher-images, and then rank these distances and return top-$K$ results. On the basis of unsupervised-learning model, our supervised-learning model seems straightforward, and we will further explore it in the future such as combining Triplet loss \cite{SchroffKP15} and Center loss \cite{WenZL016} to learn more discriminative representations. As shown in Algorithm \ref{algorithm_EViT}, we present the main processes of EViT.

    \IncMargin{1em}
	\begin{algorithm}[]
		\caption{EViT’s main algorithm}
		\label{algorithm_EViT}
		\small
		\SetKwInOut{Input}{input}
		\Input {plain-images, secret keys, labels=None} 
		\BlankLine 
		\tcp*[h] {first module: image encryption algorithm (Section \ref{proposed_encryption})} \;
        cipher-images $\leftarrow$ Image encryption (plain-images, secret keys) \;
        \tcp*[h] {second module: feature extraction method (Section \ref{feature_extraction})} \;
        features $\leftarrow$ feature extraction (cipher-images) \;
        \tcp*[h] {third module: retrieval models} \;
        unsupervised $\leftarrow$ retrieval model(features) \tcp*[h]{Section \ref{module_3}}\;
        \eIf {labels is Not None}
            {supervised  $\leftarrow$ Fine-Tuning (unsupervised, features, labels) \tcp*[h]{Section \ref{supervised}}\;
            return supervised}{return unsupervised}
	\end{algorithm}
	\DecMargin{1em}

\section{Experiments And Analysis}
	\label{experiments}
	
	In this section, experimental results of our proposed scheme are presented. We evaluate the performance on Corel10K dataset \cite{LiW03}, which is the widely used dataset by many related researches. Corel10K dataset contains 10000 images in 100 categories, with 100 images in each category. The image sizes are $128 \times 192$ or $192 \times 128$. Our programming language is Python. In the following section, we first describe the retrieval performance, then analyze time consumption of searching, finally present the encryption performance of our EViT.
	
	\begin{table}[]
		\centering
		\captionsetup{font=footnotesize}
		\caption{Descriptions of Corel10K-a and Corel10K-b datasets.}
		\label{dataset}
		\begin{tabular}{@{}lcc@{}}
			\toprule
			Datasets                & Corel10K-a & Corel10K-b \\ \midrule
			Traing set              & 7000       & 7000       \\ \midrule
			Testing set             & 3000       & 3000       \\ \midrule
			Classes of Training set & 70         & 100        \\ \midrule
			Classes of Testing set  & 30         & 100        \\ \bottomrule
		\end{tabular}
	\end{table}
	
	\subsection{Retrieval Performance}
	\label{retrieval_performance}
	
	Our EViT respectively proposes the unsupervised-learning and supervised-learning retrieval models. We compare the retrieval performance of EViT with current image-encryption-based schemes whose retrieval models are divided into unsupervised and supervised model. In order to better compare retrieval performance, we split training set and testing set with two different types, and denote as Corel10K-a and Corel10K-b datasets which are described in Tab. \ref{dataset}. For Corel10K-a dataset, we train the retrieval model on 70 classes, so testing set has no same classes with training set (open-set classification \cite{DengGXZ19}). For Corel10K-b, we train the retrieval model on 100 classes, and each class we select 70 images, and the remain 30 images of each class are in testing set (close-set classification \cite{DengGXZ19}). We use stochastic gradient descent (SGD) as optimizer. The weight decay is $5e^{-5}$, and SGD momentum is 0.9.  We set batch size to be 14 and 35 for the unsupervised-leaning and supervised-learning model respectively. We train the retrieval models using the PyTorch framework \cite{PaszkeGMLBCKLGA19} on a machine with Nvidia RTX2080Ti 11G GPU.

	The evaluation metric of retrieval performance we use is mean Average Precision \cite{0021593} (mAP) which is widely used in many retrieval tasks. When returning top-$K$ results, mAP is calculated as follows:
	\begin{equation}
		mAP@K = \frac{1}{Q} \sum_{q=1}^{Q} AP@K(q) 
	\end{equation}
	\begin{equation}
		AP@K(q) = \frac{1}{R_{q}} \sum_{k=1}^{K} p_{q}(k)\ rel_{q}(k)
	\end{equation}
	where $Q$ is the number of query images, $R_{q}$ is the number of similar images for the query $q$, $p_{q}(k)$ is precision at rank $k$ for the query $q$, and $rel_{q}(k)$ is 1 if the rank $k$ result is similar to $q$, 0 otherwise. In this paper, we use $mAP@100$ to evaluate retrieval performance. The higher $mAP@100$, the better retrieval performance.

	\begin{table}[]
		\centering
		\captionsetup{font=footnotesize}
		\caption{Comparison of retrieval performance with current schemes.}
		\label{comparison}
		\begin{tabular}{@{}llcc@{}}
			\toprule
			\multicolumn{2}{c}{Schemes}                               & Corel10K-a & Corel10K-b \\ \midrule
			\multirow{10}{*}{Unsupervised} 
			& Xia\cite{xia2019privacy}                        & 0.378     & 0.230     \\ \cmidrule(l){2-4} 
			& Liang\cite{liang2019huffman}                   & 0.321     & 0.217     \\ \cmidrule(l){2-4} 
			& Xia\cite{XiaJLLJ22}                        & 0.383     & 0.235     \\ \cmidrule(l){2-4} 
			& Xia\cite{XiaWTXW21}                     & 0.301     & 0.190     \\ \cmidrule(l){2-4}
			& Zhang \cite{zhang2014histogram}   & 0.396    & 0.269     \\ \cmidrule(l){2-4}  
			& Li\cite{LiS19}                         & 0.410     & 0.269     \\ \cmidrule(l){2-4}
			& \textbf{Ours-unsupervised} & \textbf{0.466}     & \textbf{0.295}    \\ \midrule
			\multirow{4}{*}{Supervised}   & Cheng\cite{cheng2016markov}                     & ---        & 0.407     \\ \cmidrule(l){2-4} 
			& Feng\cite{FengLLLH21}                       & 0.423     & 0.528     \\ \cmidrule(l){2-4} 
			& \textbf{Ours-supervised}   & \textbf{0.554}     & \textbf{0.759}     \\ \bottomrule
		\end{tabular}
	\end{table}
	
	Here, we compare retrieval performance with current image-encryption-based schemes. All schemes are evaluated on the same testing set, and the results of comparison are shown in Tab. \ref{comparison}. We can see that our retrieval performance is better than other schemes. Specifically, our unsupervised-learning model can achieve 0.466 $mAP@100$, which is higher about $5.6\%$ than state-of-the-art retrieval performance on the open-set Corel10K-a; on the closed-set Corel10K-b, our unsupervised-learning model can achieve 0.295 $mAP@100$, which is higher about $2.6\%$ than state-of-the-art retrieval performance. For supervised-learning model, we can significantly improve retrieval performance than other supervised schemes. It is noted that Cheng \cite{cheng2016markov} used classification probability as representations of cipher-images, it is unsuitable on open-set Corel10K-a due to testing set has no same classes with training set. So far the image-encryption-based supervised-learning schemes are few, our supervised-learning model can be a strong baseline to explore it. Next, we describe more experimental details about the unsupervised-leaning and supervised-learning retrieval performance respectively.
	
	\subsubsection{Unsupervised retrieval performance}

	Our backbone has $L$ stacked Transformer encoders (Fig. \ref{backbone}), hence we use different $L$ values to demonstrate the retrieval performances on Corel10K-a and Corel10K-b datasets. As shown in Tab. \ref{L}, we can see that $L=6$ is more suitable for our unsupervised-learning retrieval model. 
	
	\begin{table}[]
		\centering
		\captionsetup{font=footnotesize}
		\caption{Unsupervised-learning retrieval performance with different $L$ values on Corel10K-a and Corel10K-b datasets.}
		\label{L}
		\begin{tabular}{@{}lcccc@{}}
			\toprule
			$L$                &4   & 5     & 6     & 7     \\ \midrule
			Corel10K-a (mAP@100) &0.457 & 0.462 & 0.466 & 0.465 \\ \midrule
			Corel10K-b (mAP@100) &0.289 & 0.291 & 0.295 & 0.292 \\ \bottomrule
		\end{tabular}
	\end{table}

    \begin{figure}[]
		\centering
		\includegraphics[width=0.6 \textwidth]{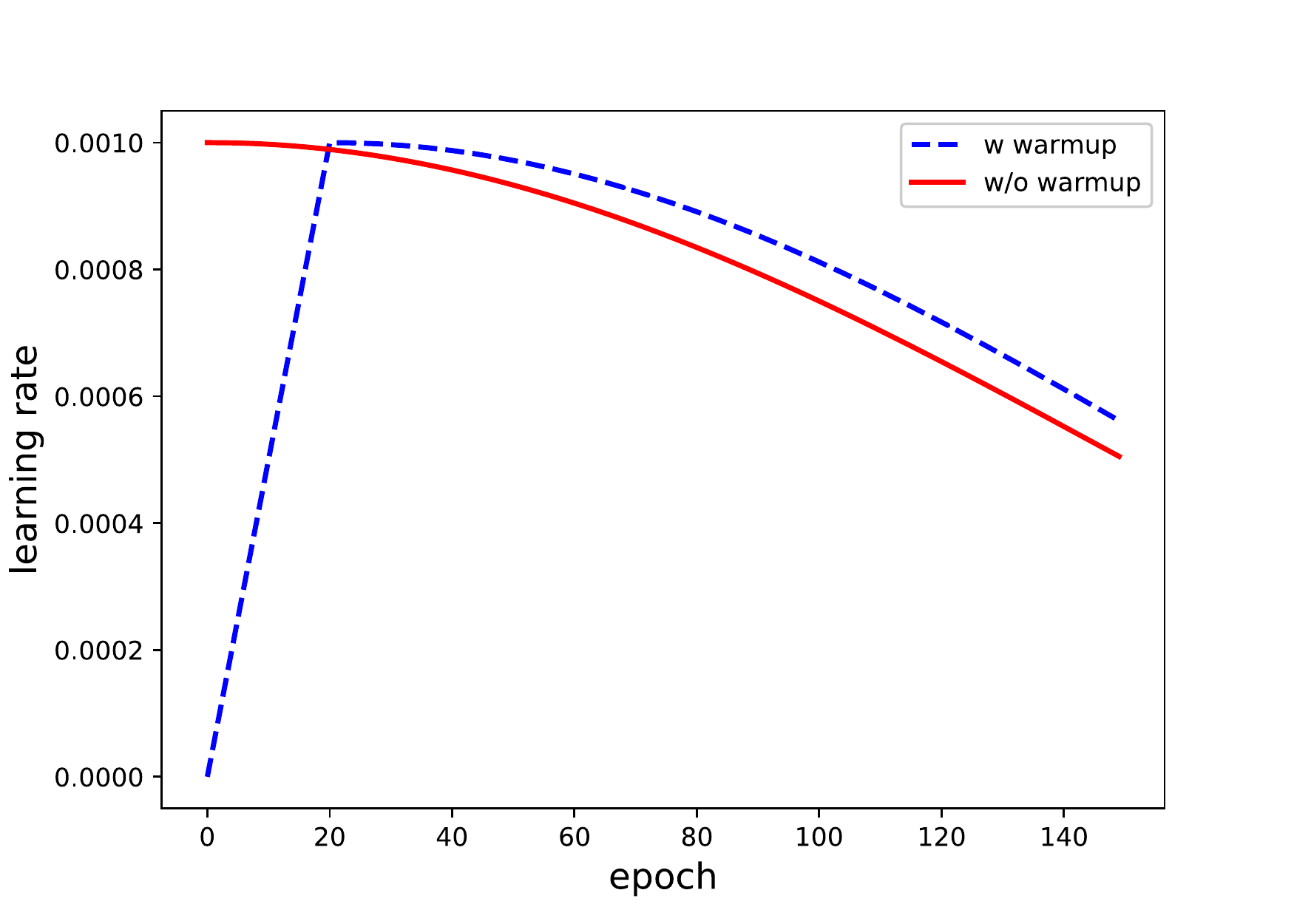}
		\captionsetup{font=footnotesize}
		\caption{Comparison of learning rate schedules.}
		\label{lr}
	\end{figure}

    We propose two adaptive data augmentations for EViT, and have mentioned that warm up and huffman embedding are helpful for our retrieval performance (Section \ref{proposed_backbone}). Here, we use ablation experiments to verify how data augmentations, warm up, and huffman embedding influence the unsupervised-learning retrieval performance on Corel10K-a and Corel10K-b datasets. Our learning rate is $1e^{-3}$ with cosine warm up, as shown in Fig. \ref{lr}, the ``blue'' line is cosine learning rate with warm up which is linearly increased to $1e^{-3}$ in the first 20 epoch; the ``red'' line is cosine learning rate without warm up. We add warm up and huffman embedding one by one ($L=6$), the results of ablation experiments are shown in Tab. \ref{ablation}. We can see that if we do not add data augmentations, warm up, and huffman embedding, the retrieval performance only can achieve 0.397 and 0.243 $mAP@100$ on Corel10K-a and Corel10K-b respectively. Our two adaptive data augmentations are helpful to enhance retrieval performance, which can improve about $1\%$ $mAP@100$. Warm up improves retrieval performance with $1.2\%$ and $1.5\%$  on Corel10K-a and Corel10K-b respectively. Huffman embedding significantly improves retrieval performance with $4.3\%$ and $2.3\%$ on Corel10K-a and Corel10K-b respectively. The huffman embedding is learned from global Huffman-code frequency which is one of our multi-level features. The ablation experiments prove that multi-level features express more abundant information of cipher-images, which can directly improve retrieval performance.

	\begin{table}[]
		\centering
		\captionsetup{font=footnotesize}
		\caption{Ablation experiments with warm up and huffman embedding for unsupervised-learning retrieval model.}
		\label{ablation}
		\begin{tabular}{@{}lcccc@{}}
			\toprule
            Data augmentations &   & \checkmark &  \checkmark & \checkmark   \\ \midrule
			Warm up      &        &      & \checkmark      & \checkmark      \\ \midrule
			Huffman embedding  &   &       &     &  \checkmark     \\ \midrule
			Corel10K-a (mAP@100) & 0.397 & 0.411 & 0.423 & 0.466 \\ \midrule
			Corel10K-b (mAP@100) & 0.243 & 0.257 & 0.272 & 0.295 \\ \bottomrule
		\end{tabular}
	\end{table}
	
	\subsubsection{Supervised retrieval performance}
	
	Our supervised model is Fine-Tuning on the unsupervised model. For example, when training the supervised-learning model on Corel10K-a dataset, our backbone can use unsupervised-learning model's parameters which are also trained on Corel10K-a as initial parameters. Unsupervised-learning model as pre-trained model for supervised-learning is common \cite{GansbekeVGPG20, ChenK0H20, He0WXG20} which can accelerate model convergence and achieve better performance. Due to the backbone of supervised-learning is same as unsupervised-learning model, most of strategies such as warm up and data augmentations also are used in supervised-learning model. Apart from loss function, the supervised-learning model is almost inspired by our unsupervised-learning model. 
	
	\begin{table}[]
		\centering
		\captionsetup{font=footnotesize}
		\caption{Retrieval performance with different $\alpha$ and $s$ on Corel10K-a/b datasets.}
		\label{supervised_parameter}
		\begin{tabular}{@{}lccc@{}}
			\toprule
			\diagbox{$\alpha$}{s} & 16    & 32    & 64    \\ \midrule
			0.1                  & 0.539 / 0.690 & \textbf{0.554} / 0.658 & 0.540 / 0.588 \\ \midrule
			0.2                  & 0.526 / 0.721 & 0.533 / 0.681 & 0.546 / 0.620 \\ \midrule
			0.3                  & 0.529 / 0.746 & 0.501 / 0.725 & 0.534 / 0.453 \\ \midrule
			0.4                  & 0.527 / 0.747 & 0.497 / \textbf{0.759} & 0.518 / 0.520 \\ \bottomrule
		\end{tabular}
	\end{table}

\begin{figure*}[]
		\centering
		\captionsetup{font=footnotesize}
		\subcaptionbox{Cheng \cite{cheng2016markov}\label{}}
		{\includegraphics[width=0.32\linewidth]{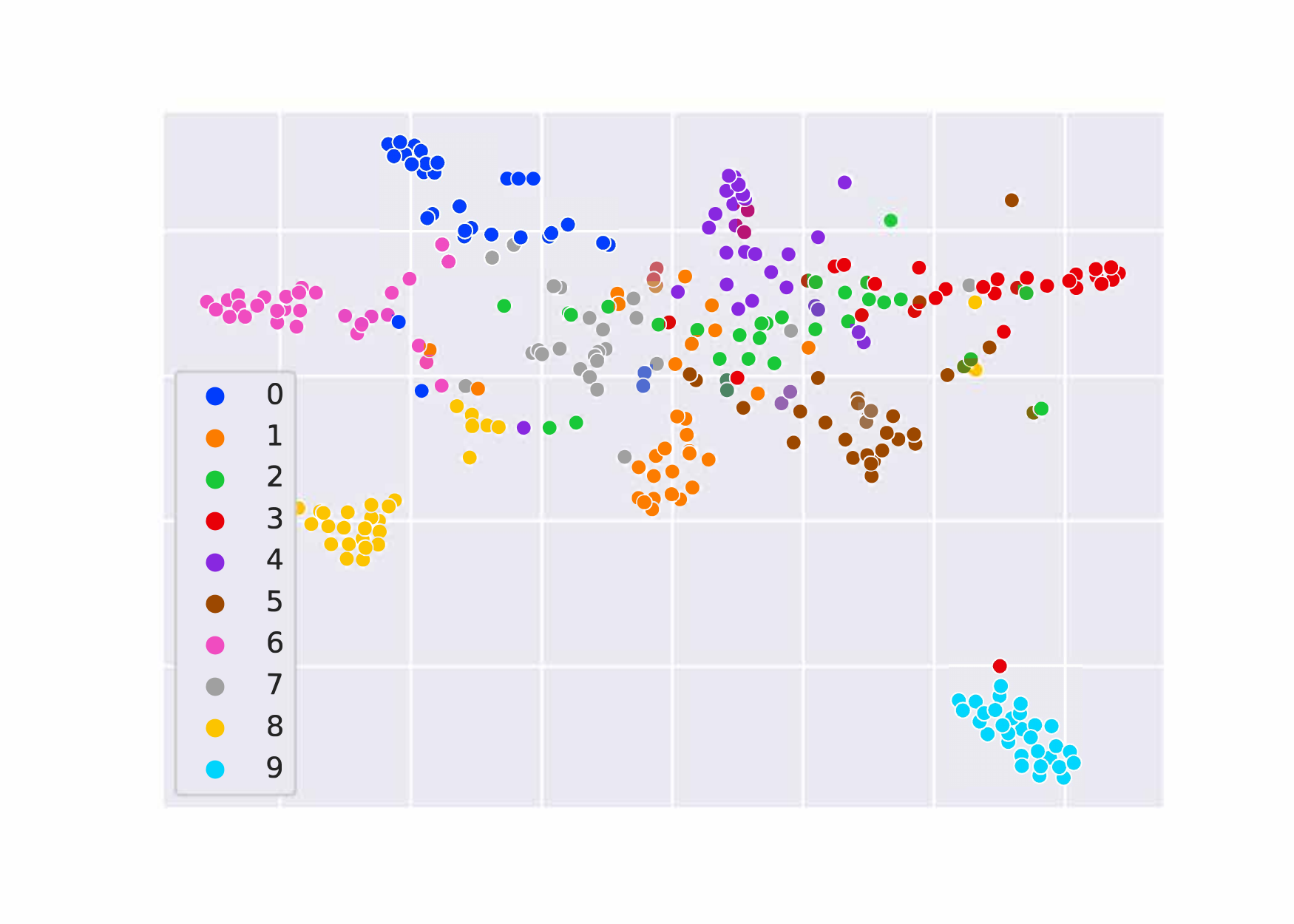}}
		\subcaptionbox{Feng \cite{FengLLLH21}\label{}}
		{\includegraphics[width=0.32\linewidth]{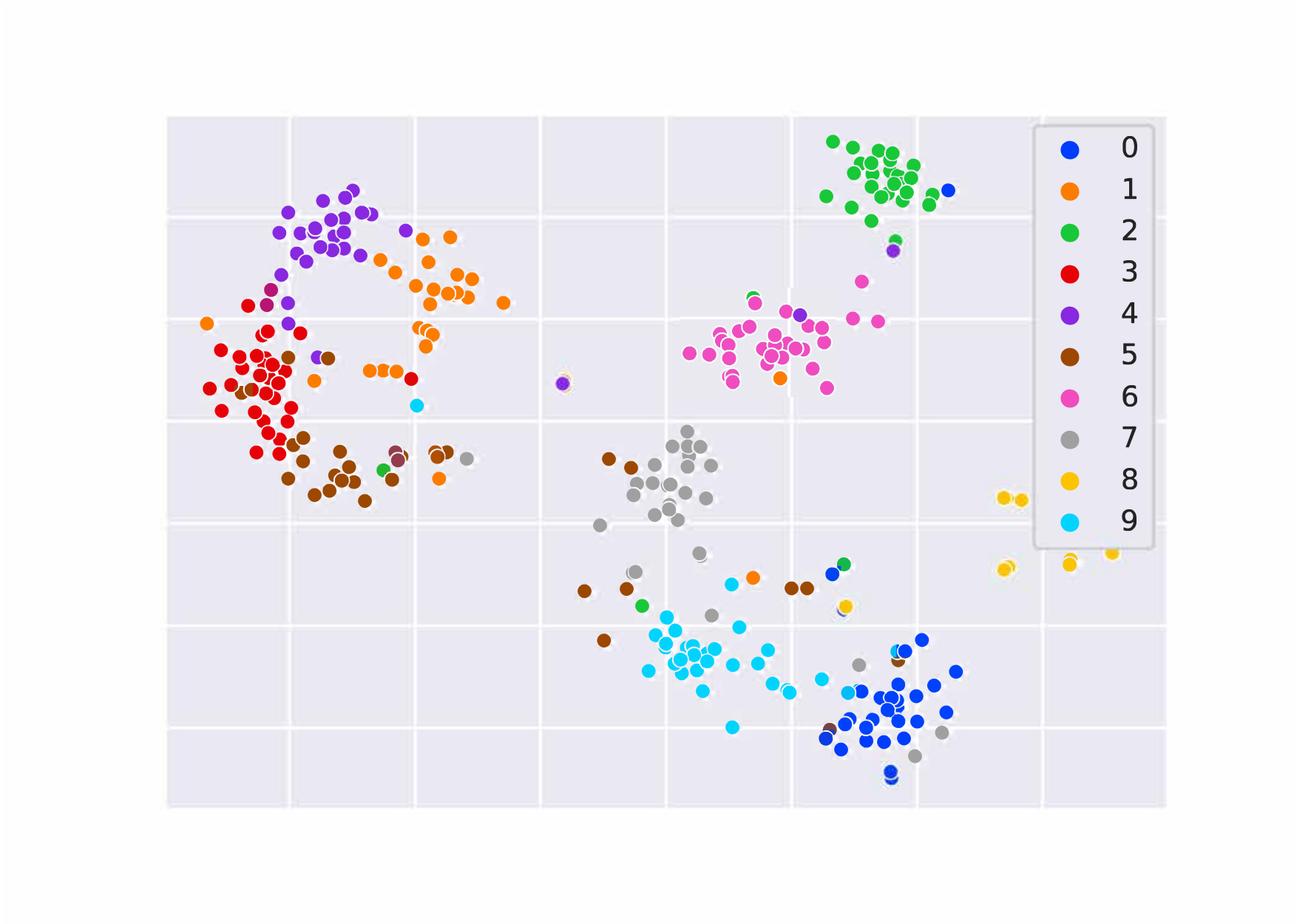}}
		\subcaptionbox{Ours\label{}}
		{\includegraphics[width=0.32\linewidth]{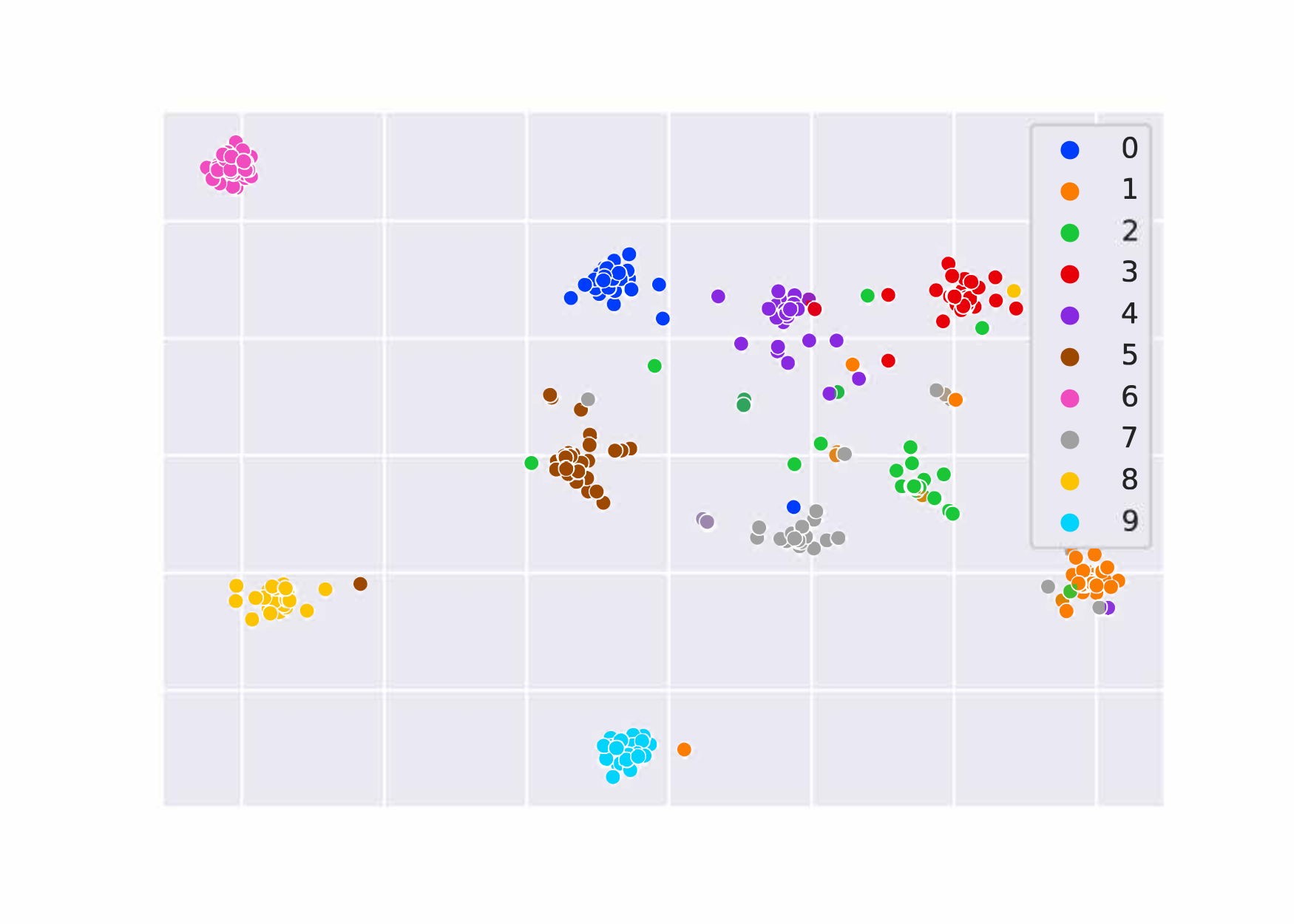}}
		\caption{Comparison of semantic space visualization with t-SNE on Corel10K-b dataset.}
		\label{tsne}
	\end{figure*}

	We mentioned that there are two hyper-parameters in ArcFace \cite{DengGXZ19}: $s$ and $\alpha$ (Eq. \ref{eq:arcface}), and now we use different $s$ and $\alpha$ to observe their influence on retrieval performance when $L=6$ on Corel10K-a and Corel10K-b (Corel10K-a/b) datasets. As shown in Tab. \ref{supervised_parameter}, we can see that on open-set dataset Corel10K-a, small $\alpha$ is more fit to supervised-learning model; while close-set dataset Corel10K-b, $s$ should not set to be too large such as 64 which will decrease the retrieval performance. Due to there are same classes in training set and testing set on Corel10K-b dataset, supervised-learning model is more easy to learn the representations. Here, we use t-SNE \cite{van2008visualizing} to visualize the semantic space of different supervised-learning schemes \cite{cheng2016markov, FengLLLH21} on Corel10K-b dataset. Specifically, 10 classes are chosen from testing set, with 30 instances in each class. Fig. \ref{tsne} shows that the semantic space of Cheng \cite{cheng2016markov} is a few disordered. Although the semantic space of Feng \cite{FengLLLH21} is separated between inter-classes, it is not compact among intra-classes. Moreover, some classes are not pulled apart. For our EViT, it is not only more distinguishable in the inter-classes but also  more closer in the intra-classes. In summary, our supervised-learning retrieval model can significantly improve retrieval performance, and there may be still room for improvement (Section \ref{supervised}). We also hope our supervised-learning model can be strong baseline to explore privacy-preserving image retrieval. 
	
    \subsubsection{Why not CNN and non-end-to-end}
	Current deep plain-image retrieval works \cite{abs-2101-11282} are end-to-end, which directly use images as model’s inputs to automatically extract features (e.g. CNN features) rather than hand-craft features. However, it is impossible for our cipher-images to extract ruled features because the spatial structure information of cipher-images (e.g., pixel values) are randomly changed and disordered by secret keys. Therefore, EViT adopts the non-end-to-end manner due to the artificial features (e.g., local length sequence and global Huffman-code frequency) are ruled and can be used to effectively learn representations of cipher-images. In Section \ref{ViT}, we have mentioned that our backbone uses ViT rather than CNN, and here we compare retrieval performance with ResNet50 \cite{HeZRS16} (a classical CNN backbone) on the Corel10K-b dataset. As shown in Tab. \ref{different_backbone}, the non-end-to-end manner is far beyond the end-to-end manner for different backbones in retrieval performance, and ViT surpasses ResNet50 about $10\%$ $mAP@100$ in the non-end-to-end unsupervised manner. Tab. \ref{different_backbone} presents that non-end-to-end manner and ViT are vital for improving retrieval performance.

     \begin{table}[]
        \centering
		\captionsetup{font=footnotesize}
		\caption{Retrieval performance ($mAP@100$) with different backbone in end-to-end (ETE) and non-end-to-end (NETE) manner on Corel10K-b dataset, “failed” represents mAP@100 less than 0.1.}
		\label{different_backbone}
        \begin{tabular}{@{}lcccc@{}}
        \toprule
        Backbone     & ResNet50 (ETE) & ViT (ETE) & ResNet50 (NETE) & ViT (NETE) \\ \midrule
        Unsupervised & failed                & failed           & 0.196                     & \textbf{0.295}                \\ \midrule
        Supervised   & 0.102                 & 0.156            & 0.598                     & \textbf{0.759}                \\ \bottomrule
        \end{tabular}
    \end{table}
	
	\subsection{Time consumption}
	
	We test the time consumption of searching, and average the results on the entire Corel10K dataset. When searching similar cipher-images, we consider that the retrieval model is already trained. As shown in Tab. \ref{PSNR_searching}, our time consumption of searching is $0.09$s which is acceptable. Because our unsupervised-learning and supervised-learning model use the same backbone, the search times are also same. For a fair comparison, all schemes use Python programming language, and Tab. \ref{PSNR_searching} shows that EViT is more faster than other schemes in term of searching. In the future, we will deploy deep retrieval model with C++ and use Faiss \cite{johnson2019billion} to further decrease searching time.

	\begin{table}[]
		\centering
		\captionsetup{font=footnotesize}
		\caption{Comparison of searching time and PSNR for different schemes.}
		\label{PSNR_searching}
		\begin{tabular}{@{}lcccccccc@{}}
        \toprule
        Schemes            & Zhang \cite{zhang2014histogram} & Liang \cite{liang2019huffman} & Xia \cite{xia2019privacy}  & Li \cite{LiS19}   & Xia \cite{XiaWTXW21}  & Xia \cite{XiaJLLJ22}  & Feng \cite{FengLLLH21} & Ours  \\ \midrule
        Searching time (s) & 1.01  & 0.11  & 0.12  & 1.01  & 0.36  & 0.35  & 0.15  & 0.09  \\ \midrule
        PSNR (dB)          & 16.40 & 9.81  & 10.17 & 13.18 & 13.12 & 13.21 & 13.65 & 12.01 \\ \bottomrule
        \end{tabular}
	\end{table}

	\subsection{Encryption performance}
	
\begin{figure*}[]
		\centering
		\includegraphics[width=0.8 \textwidth]{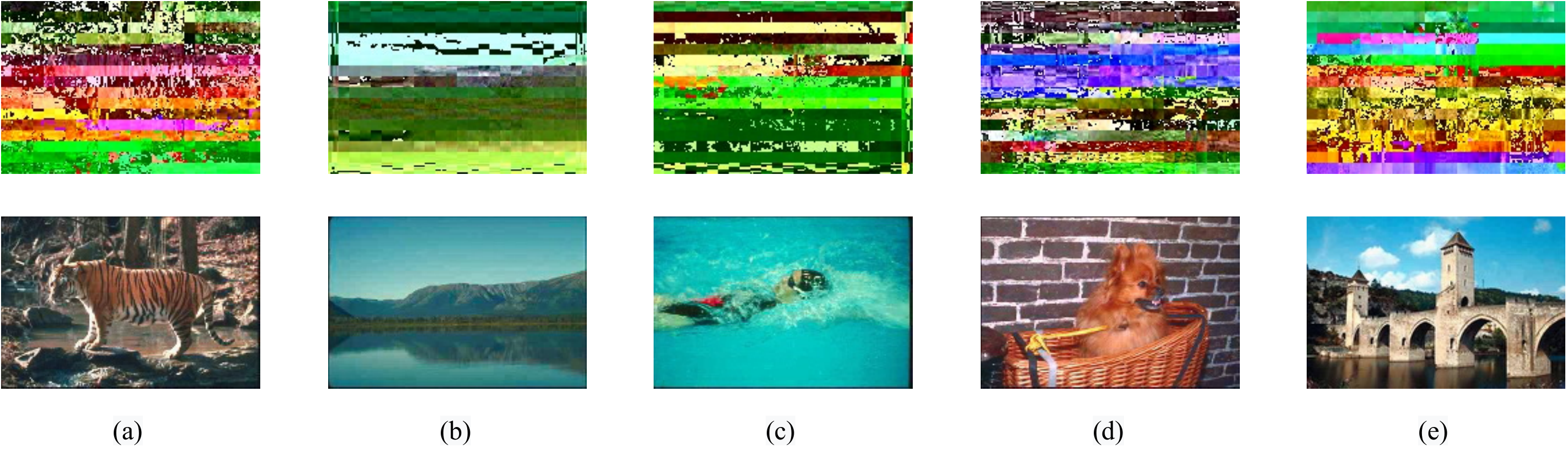}
		\captionsetup{font=footnotesize}
		\caption{Encryption examples (the first row is cipher-images, the second row is corresponding plain-images).}
		\label{encryption_examples}
	\end{figure*}

	In the proposed EViT, images are encrypted during JPEG compression. The adopted encryption operations do not destroy the format information of JPEG, hence our encryption scheme is format-compliant to JPEG. In Fig. \ref{encryption_examples}, we take five plain-images as examples to demonstrate the encryption performance of our encryption algorithm. Here, we analyze the encryption performance from four aspects: visual security, statistical attack, differential cryptanalysis, and key security. 
	
	\textit{Visual security}: According to the example shown in Fig. \ref{encryption_examples}, we can find that the encrypted images are disordered enough, and do not disclose any information of plain-images. Besides the visual checking, we use Peak Signal-to-Noise Ratio (PSNR) to evaluate the visual safety. We compare our encryption algorithm with that of current privacy-preserving image retrieval schemes, and calculate the average PSNR of 10000 images, where smaller PSNR indicates better visual safety \cite{LiL18}. All schemes are evaluated in YUV color space. As shown in Tab. \ref{PSNR_searching}, we can see that some schemes \cite{liang2019huffman, xia2019privacy} are with smaller PSNR than ours, because they used extra encryption steps apart from stream cipher. But our retrieval performance is significantly improved than them (Tab. \ref{comparison}).

\begin{figure}[]
		\centering
		\captionsetup{font=footnotesize}
		\subcaptionbox{Plain-image\label{histogram:plain}}
		{\includegraphics[width=0.45\linewidth]{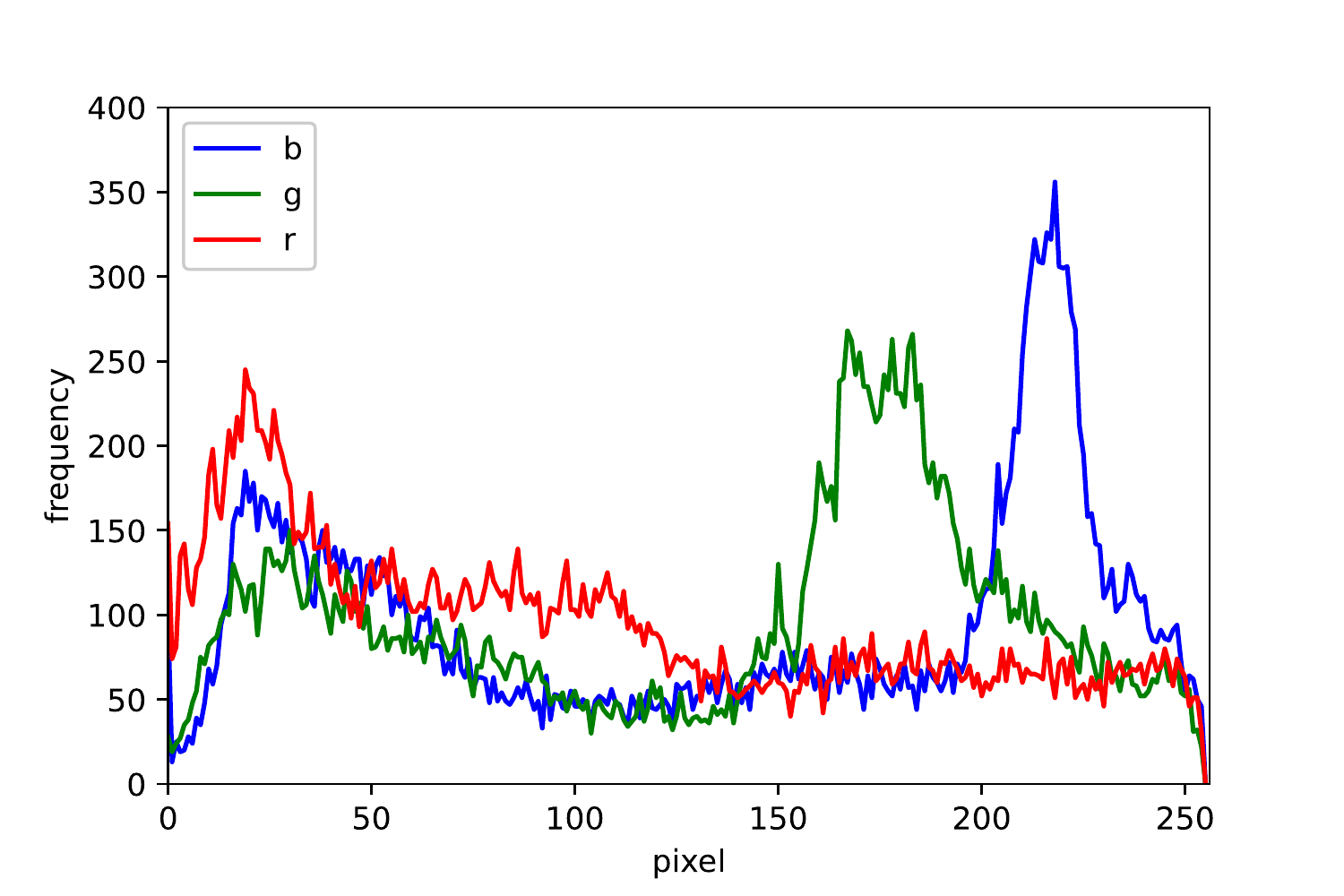}}
		\subcaptionbox{Cipher-image\label{histogram:cipher}}
		{\includegraphics[width=0.45\linewidth]{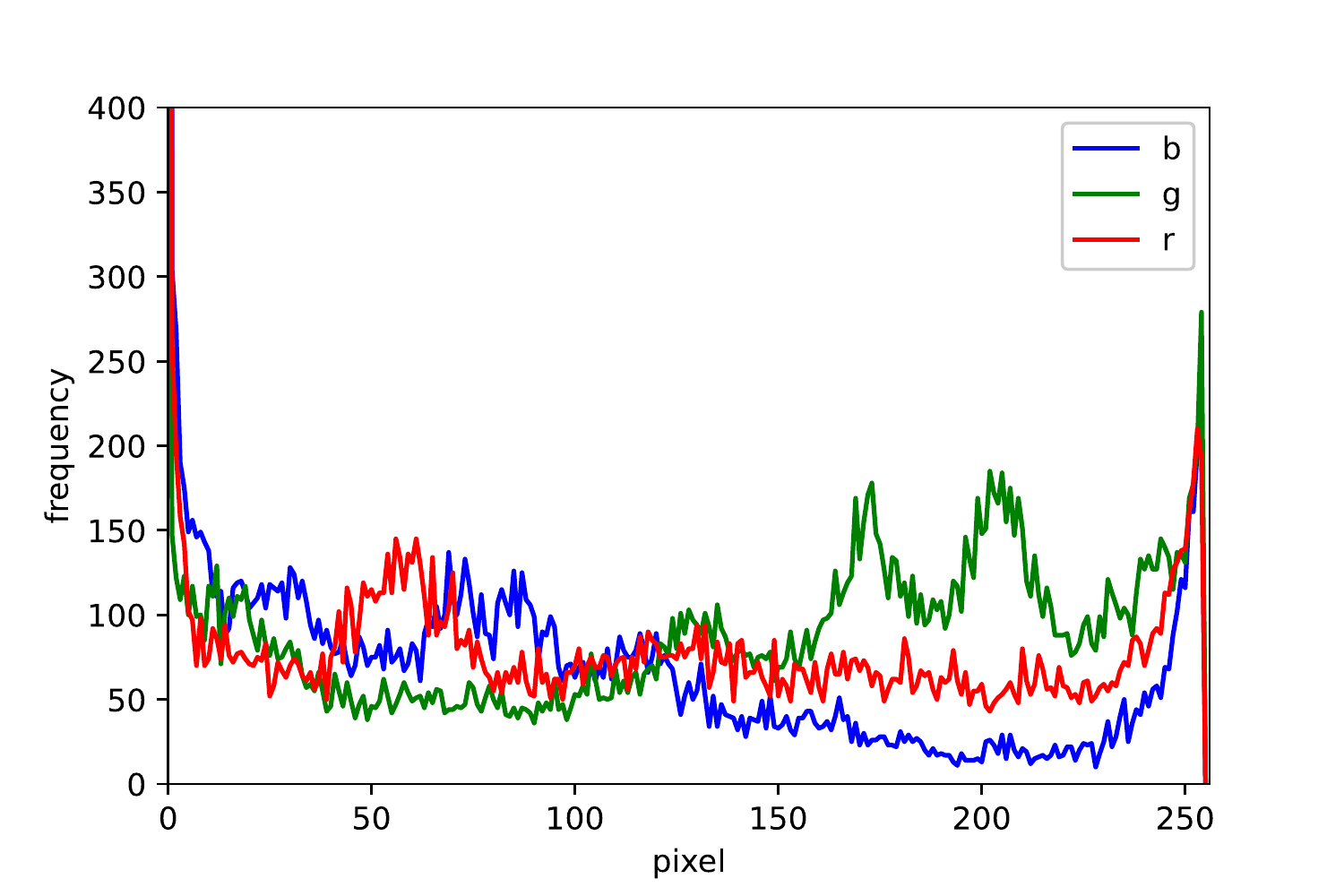}}
		\caption{Histograms of the plain and cipher images.}
		\label{histogram}
	\end{figure}

	\textit{Statistical attack}: To make the statistical mode-based attack unavailable, the histograms of plain-images and that of cipher-images should be different. As shown in Fig. \ref{histogram}, taking the plain-image (Fig. \ref{encryption_examples} (e)) and its cipher-image as examples, it can be seen that there is no statistical correlation between the histogram of the encrypted image and that of the plain-image. Compared with the histogram of the plain-image, the frequency difference of different pixel values in the cipher-image is not so large, thus our scheme has a certain resistance ability against statistical attack. 
	
	\textit{Differential cryptanalysis}: In order to resist differential cryptanalysis, minor change of plain-image such as modifying one single pixel should result in significant change in corresponding cipher-image \cite{HeHTH18}. For example, given plain-image $P^{1}$, we slightly change one single pixel and obtain plain-image $P^{2}$. Then using encryption algorithm to encrypt  $P^{1}$ and $P^{2}$, we can obtain $C^{1}$ and $C^{2}$ respectively. Differential cryptanalysis attacks cryptosystem through comparing and analyzing the differences between the cipher-images $C^{1}$ and $C^{2}$, therefore we hope there exists significant differences between them. Generally, NPCR (number of pixels change rate) and UACI (unified average changing intensity) \cite{chen2004symmetric} are two commonly metrics for evaluating the ability of encryption algorithm against differential attack, which are calculated as follows:
	\begin{equation}
		D(i,j)=
		\begin{cases} 
			0, C^{1}(i,j)=C^{2}(i,j) \\
			1, C^{1}(i,j)\neq C^{2}(i,j)
		\end{cases}
	\end{equation}
	\begin{equation}
		NPCR:N\left(C^1,C^2\right)=\frac{\sum_{i,j}{D(i,j)}}{H\times W}\times100%
	\end{equation}
	\begin{equation}
		UACI:U\left(C^1,C^2\right)=\frac{\sum_{i,j}\frac{\left|C^1(i,j)-C^2(i,j)\right|}{255}}{H\times W}\times100%
	\end{equation}
	where $C^1$ and $C^2$ are corresponding cipher-images of plain-images $P^{1}$ and $P^{2}$, and $C(i,j)$ is the pixel value at coordinates $(i, j)$ ($1\leq i\leq H, 1\leq j\leq W $). Here, we select 6 categories (church, girl, sky,architecture, painting, Africa) which contain 600 images from Corel10K, and change one single pixel of plain-image to calculate the NPCR and UACI. Each category averages the results. The closer the NPCR is to 100\% and the UACI is to 33\%, the more vital ability to resist differential attack \cite{chen2004symmetric}. As shown in Tab. \ref{NPCR_UACI}, we can see that our encryption algorithm has a certain resistance to differential attack.
	
	\begin{table}[]
		\centering
		\captionsetup{font=footnotesize}
		\caption{Mean NPCR and UACI of cipher-images with one pixel changing.}
		\label{NPCR_UACI}
		\begin{tabular}{@{}lcccccc@{}}
				\toprule
				Category & Church & Girl   & Sky    & Architecture & Painting & Africa \\ \midrule
				NPCR     & 95.55\% & 96.09\% & 96.51\% & 96.78\%       & 96.41\%   & 95.84\% \\ \midrule
				UACI     & 47.28\% & 48.13\% & 47.52\% & 48.00\%       & 48.48\%   & 48.74\% \\ \bottomrule
		\end{tabular}
	\end{table}

	\textit{Key security}: Brute-force attack is a standard ciphertext-only attack strategy in which attackers only have access to the encrypted data. Our encryption method has six encryption keys (${k_{DC}}_{*}, {k_{AC}}_{*}, * \in \{Y,U,V\}$), and each component has different keys. Each of these keys are 256-bits, so the key spaces of our encryption algorithm is $(2^{256})^{6}$. It is extremely difficult to use brute-force cracking to restore the plain images. Therefore, our scheme is safe and can be secure against ciphertext-only attack.

	EViT uses adaptive encryption key generation \cite{HeHTH18} method to encrypt images, namely different images are encrypted by different keys. We have explained the reason why EViT can use adaptive encryption key generation in Section \ref{feature_extraction}, since different keys would generate the same features of cipher-images in EViT. Some schemes such as Zhang \cite{zhang2014histogram} and Li \cite{LiS19} also can use different keys to encrypt different images, because they calculated the histograms of DCT in each frequency whose positions are unchanged, hence the features of cipher-images are unchanged. But some schemes such as Liang \cite{liang2019huffman} and Xia \cite{XiaJLLJ22} could only use the same key to encrypt different images. Liang \cite{liang2019huffman} permuted the AC coefficients in each block, so different keys will change the $(r,v)$ pairs which lead to the features extracted from cipher-images changed. Xia \cite{XiaJLLJ22} used color value replacement to encrypt images, different keys will generate different replacement tables, so the features of cipher-images also were changed. Other schemes \cite{xia2019privacy, XiaWTXW21, FengLLLH21} have the similar situations. Once the features are changed with different keys, the feature spaces of all cipher-images are different, so it leads to compromised retrieval performance.

	Our encryption algorithm cannot achieve absolute secure, and its visual safety is slightly lower than other schemes. Absolute secure encryption algorithm cannot make us extract ruled features from cipher-images, and current image-encryption-based schemes generally fail to ensure absolute security. For example, Zhang \cite{zhang2014histogram} leaked DCT histograms of cipher-images. Liang \cite{liang2019huffman} and Xia \cite{xia2019privacy} achieved smaller PSNR with visual safety than ours, but their retrieval performance is about $10\%$ $mAP@100$ lower than ours. Moreover, they used the same keys to encrypt all images, so they fail to resist differential attack. Our EViT can achieve the best retrieval performance and nice visual security among different schemes. Current privacy-preserving image retrieval schemes have been seeking balances among different performances, for example slightly compromising visual security to significantly improve retrieval performance. This is also the purpose of our work.

	\section{Conclusion}
	\label{conclusion}
	
	In this paper, we propose a novel privacy-preserving image retrieval scheme named EViT, which can improve retrieval performance by large margins than current schemes and effectively protect security of images. First, we design multi-level features (local length sequence and global Huffman-code frequency) from cipher-images which are encrypted by stream cipher with VLI code during JPEG compression process, and EViT supports adaptive encryption keys due to the the multi-level features are unchanged with different secret keys. Second, EViT proposes the unsupervised retrieval model in a self-supervised learning manner, and adopts the structure of ViT as the model's backbone to couple with the multi-level features. To improve retrieval performance, EViT adopts two adaptive data augmentations for retrieval model, and advances ViT with learnable global Huffman-code frequency. The supervised model can be easily achieved by Fine-Tuning on the trained unsupervised retrieval model. Experimental results show that EViT not only effectively protects image privacy but also significantly improves retrieval performance than current schemes. In future work, we will try to further improve the encryption performance while keeping retrieval accuracy.

	\section*{Acknowledgment}
	
	The authors would like to thank the anonymous reviewers for their constructive comments and suggestions.

\bibliographystyle{ACM-Reference-Format}
\bibliography{sample-base}
  
\end{document}